%% file: paper.tex
\documentclass[]{fairmeta}
\usepackage{makecell}
\usepackage{xcolor}
\usepackage{mathtools}

\title{Branch-Train-MiX: \\Mixing Expert LLMs into a Mixture-of-Experts LLM
}

\author{Sainbayar Sukhbaatar}
\author{Olga Golovneva}
\author{Vasu Sharma}
\author{Hu Xu}
\author{Xi Victoria Lin}
\author{Baptiste Rozière}
\author{Jacob Kahn}
\author{Daniel Li}
\author{Wen-tau Yih}
\author{Jason Weston}
\author{Xian Li}

\affiliation{FAIR at Meta}

\abstract{
We investigate efficient methods for training Large Language Models (LLMs) to possess capabilities in multiple specialized domains, such as coding, math reasoning and world knowledge. Our method, named Branch-Train-MiX (BTX), starts from a seed model, which is branched to train experts in embarrassingly parallel fashion with high throughput and reduced communication cost. After individual experts are asynchronously trained, BTX brings together their feedforward parameters as experts in  Mixture-of-Expert (MoE) layers and averages the remaining parameters, followed by an MoE-finetuning stage to learn token-level routing. BTX generalizes two special cases, the Branch-Train-Merge method, which does not have the MoE finetuning stage to learn routing, and sparse upcycling, which omits the stage of training experts asynchronously. Compared to alternative approaches, BTX achieves the best accuracy-efficiency tradeoff.
}

\date{\today}
\correspondence{\{sainbar,xianl\}@meta.com}

\begin{document}

\maketitle

\section{Introduction}
\label{section:intro}
In recent years, Large Language Models (LLMs) have shown impressive performance in a wide-range of tasks
\citep{Brown2020LanguageMA,touvron2023llama2,achiam2023gpt}, including code generation \citep{Li2022CompetitionlevelCG,Rozire2023CodeLO}, solving math problems \citep{Azerbayev2023LlemmaAO}, multilinguality \citep{zhao2024llama}, etc.
Training such LLMs requires a large amount of compute and data, exceeding thousands of GPUs and trillions of tokens.
The training parallelization is typically done by maintaining multiple copies of the model on different GPUs and keeping them synchronized after each weight update.
The cost of this frequent communication is the main bottleneck in scaling the training to more GPUs.
Besides this issue, synchronized training is more vulnerable to hardware failures as a single failed GPU can cause the whole training to halt \citep{Zhang2022OPTOP,team2023gemini}.

Recent work by \cite{Li2022BranchTrainMergeEP} proposed the Branch-Train-Merge (BTM) method for embarrassingly parallel training of LLMs 
without any synchronization for improving the throughput of pretraining.
It starts by creating multiple copies of a seed LLM, then separately training each copy on different subsets of data.
This results in multiple independent LLMs that do not share any parameters and each LLM is an expert specializing in its own data distribution, such as knowledge domains, languages or even modalities.
At test time, an input prompt is classified into one or more of the domains, and then the final outputs are formed from the corresponding expert models which are combined to predict the next token.
While this approach makes training more efficient, its main drawback is the lack of a unified single model making it impossible to do further supervised finetuning (SFT) or reinforcement learning from human feedback (RLHF) finetuning \citep{Ouyang2022TrainingLM}, both of  which can boost performance further, and are 
crucial steps in building aligned LLMs. 

A separate line of work for reducing the computational footprint of LLMs is the Mixture-of-Experts (MoE) approach \citep{Jacobs1991AdaptiveMO,Shazeer2017OutrageouslyLN}, where only a subset of parameteters are active at any given time.
In particular, MoE is applied to the feedforward sublayer of Transformers \citep{fedus2022switch,Roller2021HashLF,Lewis2021BASELS}, allowing the total number of parameters to grow without additional computation.
LLMs scaled in this way have shown impressive performance on downstream tasks \citep{Jiang2024MixtralOE, xue2024openmoe}. Unlike Branch-Train-Merge, Mixture-of-Experts are often trained in a fully synchronized fashion, and the communication cost increases with the number of experts due to all-to-all communication.

\begin{figure}
    \centering
    \includegraphics[width=0.8\linewidth]{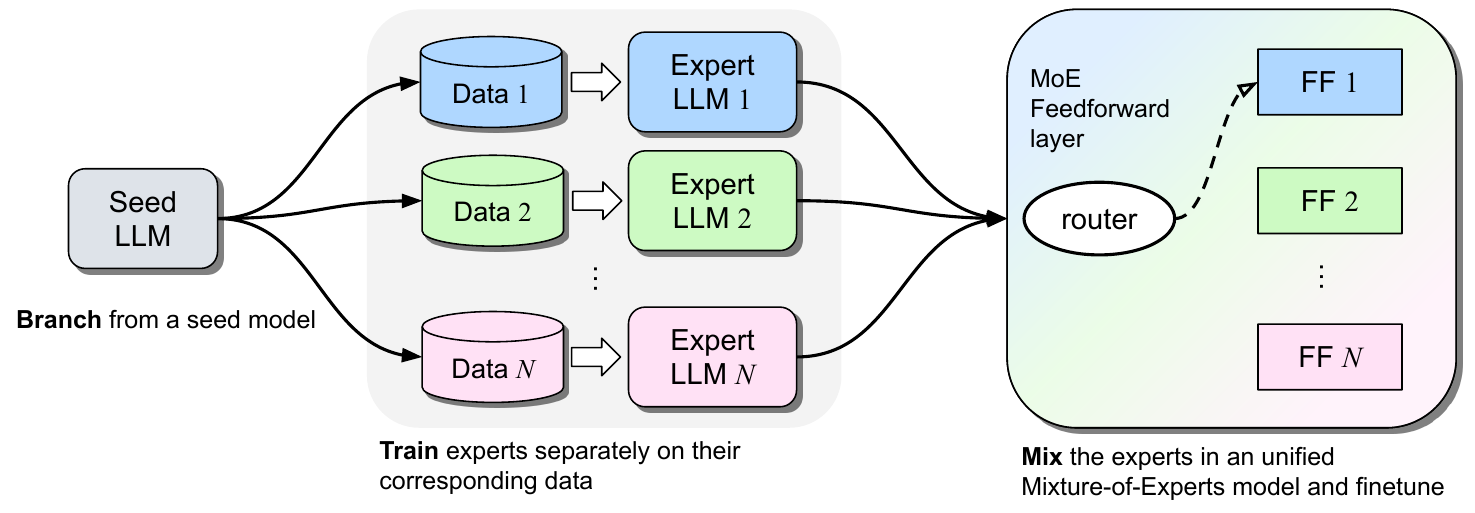}
    \caption{\textbf{The Branch-Train-MiX (BTX) method} has three steps: {\bf 1) branch} from a pretrained seed LLM by making multiple copies of it; {\bf 2) train} those copies separately on different subsets of data to obtain expert LLMs; {\bf 3) mix} those expert LLMs by combining them into a single LLM using mixture-of-experts feedforward (FF) layers, and finetuning the overall unified model.}
    \label{fig:method}
\end{figure}

In this paper, we aim for the best of both worlds, combining the advantages of Branch-Train-Merge and Mixture-of-Experts, while mitigating their disadvantages.
We achieve this by training multiple expert LLMs separately as in the Branch-Train-Merge method, but subsequently combine those experts into a single model using an MoE architecture.
More specifically, the feedforward sublayers from all the expert LLMs are brought together into a single MoE module at each layer, and a router network selects which feedforward expert to use at every token.
We merge other modules of the expert LLMs, including self-attention layers, by simply averaging their weights.
Then the resulting model is MoE-finetuned
on all the combined data by continuing training, so that the router can learn to mix the expert feedforward (FF) modules.
\autoref{fig:method} shows an overview of this method, which we call \emph{Branch-Train-MiX} (BTX).

The main advantage of BTX compared to MoE is that expert training is embarrassingly parallel and asynchronous, reducing communication cost and increasing training throughput.
Compared to Branch-Train-Merge, the final BTX model is a unified neural network that can be finetuned or used like any other standard LLM.
The final BTX model will not significantly increase inference FLOPs compared to the seed model since it is sparsely activated, despite having a much larger number of parameters.

We conduct our experiments using \textsc{Llama-2 7B} \citep{touvron2023llama2} as a seed model and train expert LLMs on different subsets of data corresponding to the domains of math, code and Wikipedia.
With the original \textsc{Llama-2 7B} weights added as a fourth expert, we finetune the combined MoE model for a relatively short period compared to the pretraining process.
The resulting BTX model brings significant improvements over the seed model on tasks across various domains, especially bridging the gap with specialized models on math and code related tasks, while retaining performance on the original capabilities where specialized models suffer from catastrophic forgetting. BTX outperforms BTM on all tasks demonstrating the benefits of learnt routing through MoE finetuning. Compared to purely MoE training such as sparse upcycling, BTX is more compute efficient with higher training throughput and more balanced performance across tasks in different domains.

\section{Related Work}

\paragraph{Asynchronous parallel training}
Reducing communication between training workers for computational efficiency is a major topic of study for training deep learning systems.
\cite{NIPS2015_d18f655c} introduced a method that allows model instances on different workers to diverge from each other, thus eliminating the constant need of synchronization. Instead, the workers are loosely synchronized to  master weights using elastic averaging from time to time.
A more recent work by \citet{Douillard2023DiLoCoDL} showed that less frequent synchronization of diverged workers by averaging their weight changes and applying Nesterov momentum works well in practice for training LLMs.
The Branch-Train-Merge method \citep{Li2022BranchTrainMergeEP,gururangan2023scaling} takes parallel training to the extreme by running multiple training processes completely independently.
Each training process uses specific domain data, thus the corresponding model becomes an expert in that domain. Finally, the output distributions of those expert models are averaged to make a next token prediction.
Which experts to average is decided by classifying the input into one or more of the domains.
\citet{Wortsman2022ModelSA} showed simply averaging parameters of separately trained models improves performance, but the models only differed in their hyperparameters.

\paragraph{Mixture-of-Experts}
MoE is used to scale deep networks in \cite{Shazeer2017OutrageouslyLN} using a simple Top-K routing scheme.
Since the routing decisions are discrete and thus cannot be trained by gradient descent, various training methods have been explored for the Transformer architecture \citep{fedus2022switch,Lewis2021BASELS}.
Surprisingly \cite{Roller2021HashLF} showed that even a fixed routing scheme without any learning works well, if the routing is done via a random mapping based on input tokens.
In larger scale experiments with recent LLMs, \cite{Jiang2024MixtralOE} demonstrated that the MoE approach can match the performance of dense LLM counterparts using a much smaller number of active parameters.
A study by \cite{Dai2024DeepSeekMoETU} showed the advantage of more fine-grained experts, as well as having a shared expert that always stay active.
More similar to our work, \cite{Gururangan2021DEMixLD} makes experts in  feedforward layers  specialize to specific domains using a domain-conditioned fixed routing, but it lacks the asynchronous training of our approach.

\paragraph{Continual learning}
Our method relates to continual learning  \citep{awasthi2019continual} because domain experts are trained on datasets with different distributions from the initial data used for training the seed model, which is implemented by continued training after branching.
Specifically, our approach is related to parameter isolation methods \citep{DeLange2019ACL} as we have different parameters for different domains.
\citet{Aljundi2016ExpertGL} also creates a new copy of a model to train on each domain.
\citet{Rusu2016ProgressiveNN} adds a new model with a new domain, but connects it to the previous models so the previously learned features can be used.
\citet{Rozire2023CodeLO} showed continual training of a seed LLM on a specific domain of code can produce a strong domain expert model, and this converges much faster than starting from scratch.
For training a math expert, starting from a code expert rather than a general LLM was shown to be more beneficial \citep{Shao2024DeepSeekMathPT,Azerbayev2023LlemmaAO}.

\section{Branch-Train-MiX}
\label{section:methods}

Given an existing LLM $\mathcal{M}$ which has been pretrained on a large corpora covering a wide variety of topics, we aim to improve its performance on $N$ areas of expertise. This is achieved by continued pretraining with corresponding training  datasets $\mathcal{D} \coloneqq \{ D_1, \ldots, D_N \}$, each related to a specific knowledge domain such as math, code, etc. 
The proposed method contains three stages: Branch, Train, and MiX. 

\subsection{Branch \& Train: Embarrassingly Parallel Expert Training}

Initializing from the seed model  $\mathcal{M}$, we train $N$ expert LLMs $\{ \mathcal{M}_1, \ldots, \mathcal{M}_N \}$, with each model $\mathcal{M}_i$ being trained on the corresponding dataset $D_i$ in the same manner as during pretraining, using the usual language modeling objective.
Since each expert model  $\mathcal{M}_i$ can be trained in complete separation from the others, the whole training process becomes $N$-way embarrassingly parallel. This training paradigm has several benefits in large-scale distributed training. It allows linear scaling of overall training throughput when scaling up the size of compute, while joint training often faces uncertain performance from increasing batch size. It has lower all-to-all communication cost. It is also more resilient, as a single training failure will only affect one of the $N$ training processes instead of halting the entire training.

After all the expert training is finished, we will end up with $N$ different LLMs, with each specializing in a specific distribution. 
At this point, the Branch-Train-Merge method \citep{Li2022BranchTrainMergeEP,gururangan2023scaling} uses these domain experts as is, choosing which expert to use by determining which domain the input belongs to at inference time.
Usually multiple experts are chosen, and their final output distributions are simply averaged to generate the next token.
Our BTX approach, in contrast, merges these domain experts back into a single LLM that is finetuned further, as we will describe in the next section.

\subsection{MiX: Combining Separate Experts to be a Mixture-of-Experts}
\label{sec:method_mix}
We employ a Mixture-of-Experts approach to combine the domain expert models $\mathcal{M}_i$.
However, instead of using the classical procedure of mixing the final outputs from $\mathcal{M}_i$, we do  a more fine-grained mixing by performing MoE within each layer of a Transformer.
In particular, we combine the different feedforward sublayers from the domain experts  into a single MoE sublayer.
If $\mathtt{FF}_i^l(x)$ is the feedforward sublayer at the $l$-th layer of the $i$-th domain expert $\mathcal{M}_i$, then the combined MoE layer for input representation $x$ at layer $l$ will compute:
\[
\mathtt{FF}_\text{MoE}^l(x) = \sum_{i=1}^N g_i(W_l x) \mathtt{FF}_i^l(x).
\]
Here $W_l$ is a linear transformation and $g$ is a routing function, which usually has sparse output and hence switches on only some experts.
Since we can skip computing $\mathtt{FF}_i^l(x)$ if the corresponding router output is zero, the actual computation of $\mathtt{FF}_\text{MoE}^l(x)$ will be much more efficient than computing all domain experts.
However, routing decisions can change from token to token, so one input sequence can employ all the domain expert FF layers if needed, even when only a few are accessed at any given token.
In our experiments, we use Top-k (k=2) routing where $g(W_l x) = \text{SoftMax}(\text{TopK}(W_l x))$, unless otherwise stated.

For the self-attention sublayers, we combine the different domain experts by simply averaging their weights. The motivation behind this is the assumption that the self-attention layers are less domain specialized than the feedforward layers.
We do the same averaging for the remaining parameters (embeddings, etc.) as well.

Note that the only new parameters we introduce are the router's transformation parameters $W_l$, which are negligible in size compared to the rest of the network.
Nevertheless, those new parameters need to be finetuned, so the router can make optimal decisions in selecting which domain $\mathtt{FF}_i$ to use.
In addition, funetuning is helpful because the self-attention weights are constructed by averaging, and are likely not optimal.
Overall, the entire system has not been optimized for working together at all in the embarrassingly parallel training framework, but our hypothesis is that even a small amount of combined finetuning might make large improvements.

\subsection{Variations}
\label{sec:variation}
We also experimented with several variations of our method.

\paragraph{Load balancing} A common problem with MoE is the emergence of dead experts, which do not get activated by the router at all.
Common routing methods like Top-k are unlikely to escape from such a situation because a dead expert is never in the top-k selection, and therefore never receives a training signal.
Load balancing offers a simple solution by adding an extra loss term that encourages the experts to be utilized equally.
We use a loss term similar to \citep{fedus2022switch}:
\[
\mathcal{L}_\text{LB} = \alpha N \sum_{i=1}^N u_i p_i \quad \text{where} \ u_i = \frac{1}{|\mathcal{B}|} \sum_{x \in \mathcal{B}} g_i(W_l x)  \ \text{and} \ p_i = \frac{1}{|\mathcal{B}|} \sum_{x \in \mathcal{B}} \text{SoftMax}_i(W_l x) .
\]
Here $\mathcal{B}$ is the current data batch, and $\alpha$ is a hyperparameter. This loss is computed in each layer and added to the NLL loss.

\paragraph{Routing method} Besides Top-k routing, we also experiment with other routing methods:
\begin{itemize}
    \item Switch: It is a Top-1 routing method proposed by \citet{fedus2022switch}.
    \item Soft routing: We use softmax as the routing function $g$, so all experts are activated both during training and inference. While it is likely to provide the best performance, it comes at the expense of increased compute.
    \item Sample Top-1: We use the gumbel softmax \citep{jang2016categorical} for $g$.
    At training time, we generate a soft sample from the gumbel softmax, but zero out all its values except the largest one. Then we compute only one expert  corresponding to this largest value, omitting the other expert computations. At inference time, we simply do hard sampling. We anneal the temperature to a sharp distribution at the end of training to gradually reduce the discrepancy between training and inference. 
\end{itemize}

\paragraph{Splitting Experts} The number of modules in the MoE layer matches the number of domains we train on, since each module corresponds to one domain.
However, we can increase the number of modules in a simple way by splitting each domain FF sublayer into multiple chunks.
Given $N$ domains and an FF activation size of $d_\text{FF}$, we split each FF layer into $C$ chunks with a dimension of $d_\text{FF}/C$.
As a result, the final MoE layer will have $MC$ modules.

\paragraph{Blending Experts} Instead of directly initializing MoE experts from domain experts in a one-to-one way, we also try including all domains in each MoE expert.
The motivation behind this is an observation that MoE experts trained in a standard way do not show domain specialization, but rather are activated uniformly across different domains \citep{Jiang2024MixtralOE}.
In contrast, our domain experts are specialized to a specific domain  through their training data.
To break this domain specialization, we split each domain expert's FF layers into $N$ chunks and then merge the $n$-th chunks from all domains to build the $n$-th MoE expert.
This way, each MoE expert contains the same amount of parameters from all domains.

\section{Experiments}

\subsection{Experimental Setup}
We base our experiments on the setup used for \textsc{Llama-2} pretraining \citep{touvron2023llama2}.
In particular, we use the \textsc{Llama-2} 7B model as our seed model.

\subsubsection{BTX Training}
\label{sec:btx_training}

We use the pretrained Llama-2  \citep{touvron2023llama2} with 7B parameters as our seed model. 
After making three copies of the seed model \textsc{Llama-2 7B}, we continue training them on the following domain datasets to derive three domain experts:

\begin{itemize}
    \item {\bf Math:} The same data sources and mixture used in Llemma \citep{Azerbayev2023LlemmaAO} model training. To be comparable to Llemma, we train on the same amount of data as well, i.e. 48k steps with 201B tokens in total. 
    \item {\bf Code:} The same data sources and mixture of code data used in \textsc{CodeLlama} pretraining \citep{Rozire2023CodeLO}.  
    The code expert LLM is trained for 50k steps with 210B tokens in total to be comparable with the math expert.
    \item {\bf Wikipedia:} Wikipedia documents extracted between June to August 2022. The data was preprocessed to remove hyperlinks, comments and other formatting boilerplate. Since this is a smaller dataset, we train a total of 42B tokens.
\end{itemize}

While we can proceed with only these three domain experts, we also include the original seed LLM as a ``generalist'' expert so that its general knowledge is transferred to the final model.
Thus we mix these four expert models into a single MoE model as described in \Cref{sec:method_mix}.
Then we finetune this MoE model on all the data sources used to train the four experts (including the original \textsc{Llama-2 7B} pretraining data for the generalist expert) and train for another 80B tokens.   
The detailed sampling ratio across datasets in each domain as well as across the domains is described in \autoref{appendix:data}.
For BTX with default Top-2 routing, we use load balancing with $\alpha=0.01$, unless otherwise stated. For the Sample Top-1 routing, we use the temperature annealing schedule $\tau$=max($0.5, -rt$) from \cite{jang2016categorical} with $r=1e-4$ where $t$ is the number of training steps. 
For the first layer only, we used soft-routing instead.
Since the Sample Top-1 training is more efficient than Top-2, with the same compute budget it can train 160B tokens. 

\begin{table*}[t]
    \centering
    \small
    \begin{tabular}{lccccccc}
    \toprule
    & \multicolumn{2}{c}{\bf Math} & \multicolumn{2}{c}{\bf Code} & \multicolumn{3}{c}{\bf General knowledge} \\
    \cmidrule(lr){2-3} \cmidrule(lr){4-5} \cmidrule(lr){6-8}
    & \bf GSM8K  & \bf MATH & \bf Human  & \bf MBPP  & \bf Natural & \bf Trivia             & \bf MMLU  \\
    &  &  & \bf Eval & & \bf Questions  & \bf QA  &  \\    
    \midrule
    \textsc{Llama-2 7B} & 14.7 & 2.5 & 12.8 & 20.8 & 16.4 & {\bf 58.5} & 46.1  \\
    Math expert & {\bf 39.5} & {\bf 18.8} & 25.0 & 33.6 & 14.4 & 37.1 & {\bf 52.0} \\
    Code expert & 12.0 & 4.0 & {\bf 31.7} & {\bf 40.2} & 11.5 & 29.9 & 39.6 \\
    Wikipedia expert & 11.7 & 3.1 & 11.0 & 15.2 & {\bf 21.8} & 57.2 & 43.1 \\
    \bottomrule
    \end{tabular}
    \caption{Individual domain expert LLM performance on representative tasks, compared to the seed model \textsc{Llama-2 7B}. As expected, the code and math experts excel at their corresponding domain tasks.
    The Wikipedia expert performs better on Natural Questions, but the math expert has the best score on MMLU.
    This could be because MMLU contains many math subjects and math training is shown to help on this task \citep{Shao2024DeepSeekMathPT}.}
    \label{table:single_expert}
\end{table*}

\subsubsection{Baselines}

We compare to the following baselines:

\begin{itemize}
    \item {\bf \textsc{Llama-2}:} We compare to the original \textsc{Llama-2 7B} that we use as a seed model, as well as \textsc{Llama-2 13B}.
    \item {\bf Dense:} Instead of training separate LLMs on different domain datasets, the dense baseline continues to train the seed LLM with all the data.
    We use exactly the same training data as BTX, first training on the new domain-specific data used in the experts training stage, followed by the same data mixture that includes the \textsc{Llama-2} pretraining data in the MoE finetuning stage.
    We call this comparison \emph{data-matching} (DM).
    \item {\bf Sparse upcycling:} 
    This baseline \citep{Komatsuzaki2022SparseUT}
    initializes a MoE model from the seed model by making 4 identical copies of the feedforward module as experts. We use the Top-2 router with randomly initialized $W_i$ parameters.
    In addition to training a data matching baseline with the same data as is used in BTX and the dense baseline, we also train a sparse upcycling baseline with the same amount of GPU-days, i.e. compute-matching (CM), using the MoE finetuning data mixture throughout training. This is equivalent to a special case of BTX which does not contain embarrassingly parallel expert training.

    \item {\bf Branch-Train-Merge (BTM):} This baseline \citep{Li2022BranchTrainMergeEP} uses the same expert LLMs as BTX (including the original seed model) but uses them directly without building a MoE model. For a given context (input), it selects Top-k expert LLMs based on the similarity between the context and experts' training data. Following the efficient inference method used in \citet{gururangan2023scaling}, both context and experts' training data are embedded via tf-idf. Top-k experts are selected based on cosine similarity to the mean tf-idf embedding of each expert. 
    \item {\bf CodeLlama 7B:} A language model specializing in code \citep{Rozire2023CodeLO} by continued training of the same seed model \textsc{Llama-2 7B} on code data. It also has other features such as long-context and infilling.
    \item {\bf Llemma 7B:} A language model  specializing in mathematics \citep{Azerbayev2023LlemmaAO} by continued training of CodeLlama 7B on math data.

\end{itemize}

We use the same optimization hyperparameters for training of the baselines, expert models and MoE models. We use the AdamW optimizer with weight decay 0.1, and anneal the learning rate to the peak of $1e-4$ with 100 steps of warmup, and decay to $10\%$  of the peak with a cosine schedule. We use a batch size of 4M tokens with a sequence length of 4096.

\begin{table*}[t]
    \centering
    \small
    \begin{tabular}{lcccccc}
    \toprule
    & \bf Math & \bf Code & \bf Knowledge & \bf Reasoning & \bf MMLU & \bf Average \\
    \midrule
    \midrule
    \footnotesize \hspace{-2mm}\textit{Specialized LLMs} &  &  &  &  &  &  \\   
    \textsc{CodeLlama 7B} & \phantom{0}8.1 &  36.3 & 22.2 & 56.6 & 38.6 & 37.9 \\
    \textsc{Llemma 7B} & 28.0 & 33.5 & 17.2 & 38.8 & 33.5 & 32.1  \\
     \midrule
    \footnotesize \hspace{-2mm}\textit{Generalist LLMs} &  &  &  &  & &   \\
    \textsc{Llama-2 7B} & \phantom{0}8.6 & 16.8 & 37.4 & 63.3 & 46.1 & 40.7  \\ 
    \textsc{Llama-2 13B} & 16.3 & 24.5 & 40.0 & {\bf 66.1} & 52.8 & 45.4  \\
    Dense (DM) & 18.3 & 25.8 & 39.6 & 63.3 & 49.8 & 44.5  \\
    Sparse upcycling (DM), Top-2 & {\bf 28.1} & 34.7 & 34.0 & 62.3 & 51.1 & 46.3  \\
    BTM, Top-1 & 21.3 & 36.4 & 26.5 & 61.0 & 44.3 & 43.1 \\
    \vspace{1mm}
    BTM, Top-2 & 21.5 & {\bf 36.6} & 26.9 & 61.2 & 44.3 & 43.4 \\
    BTX, Sample Top-1 & 26.4 & 31.5 & 40.1 & 63.7 & {\bf 53.2} & 47.3  \\ 
    BTX, Top-2 &  27.4 & 34.0 & {\bf 41.0} & 63.5 & 52.5 & {\bf 47.9}  \\
    \bottomrule
    \end{tabular}
    \caption{Aggregated performance of BTX compared against various baselines, including both generalist and specialized pretrained models, tested on various capabilities aggregated across popular benchmarks. Dense, sparse upcycling, BTM and BTX are trained on exactly the same amount and mixture of data with the exception that BTM does not have the finetuning stage.
    }
    \label{table:btx_overall}
\end{table*}

\subsubsection{Evaluation}

For evaluation, we use the zero- and few-shot performance on multiple benchmarks that test different skills: 
\begin{itemize}
    \item Math: we report the average performance on GSM8K (8 shot) \citep{cobbe2021gsm8k} and MATH (4 shot) \citep{Hendrycks2021MeasuringMP} for math reasoning.
    \item Code: we report the average performance of HumanEval (0 shot) \citep{Chen2021EvaluatingLL} and MBPP (3 shot) \citep{Austin2021ProgramSW} for code generation.
    \item World knowledge: we report the average performance of Natural Questions (5 shot)\citep{47761} and TriviaQA (5 shot) \citep{Joshi2017TriviaQAAL}.
    \item Reasoning: we report the average 0-shot performance of ARC-Easy and ARC-Challenge \citep{Clark2018ThinkYH}, SIQA \citep{sap2019socialiqa}, PIQA \citep{bisk2020piqa} and WinoGrande \citep{sakaguchi2021winogrande}.
    \item General: we report performance on MMLU (5 shot) \citep{DBLP:conf/iclr/HendrycksBBZMSS21} which covers multiple domains. 
\end{itemize}

\begin{figure*}[t]
     \centering
     \hfill
     \includegraphics[width=9cm]{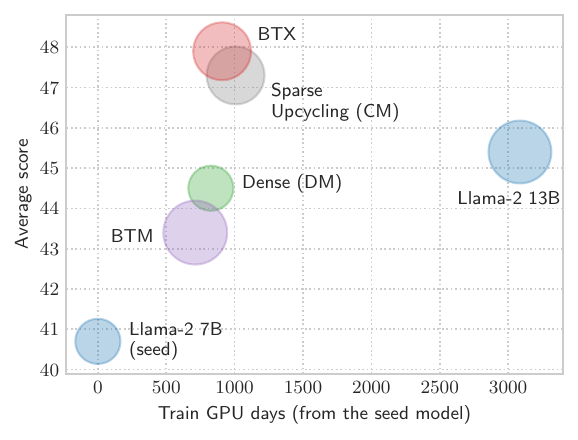}
     \hfill
     \includegraphics[width=6cm]{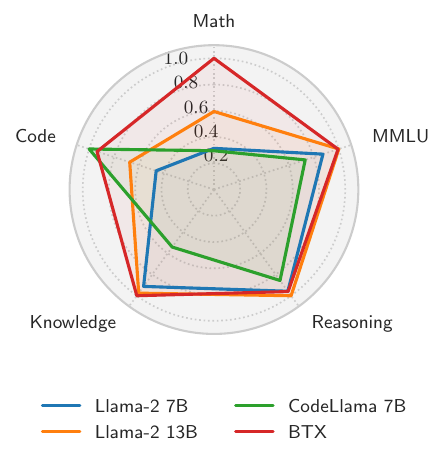}
     \hfill
     \caption[performance]{
     {\bf Left:} The average performance vs training budget of BTX compared to various baselines, with different active parameters at inference time indicated by circle size. All the models except  \textsc{Llama-2} 13B are trained starting from \textsc{Llama-2} 7B using the datasets described in \Cref{sec:btx_training}. The X-axis shows the total training compute starting from the seed model measured in GPU days\footnotemark, and the Y-axis is the average score over all the tasks (as computed in \autoref{table:btx_overall}). The BTX models outperform the baselines that started from the same seed model, as well as \textsc{Llama-2 13B}. 
     {\bf Right:} The normalized performance over different domains where the scores are divided by the highest one. We see large improvements for BTX in code (which matches the specialized model) and math tasks compared to the seed model \textsc{Llama-2 7B}, even outperforming the \textsc{Llama-2 13B} model.
     }
     \label{fig:plots}
\end{figure*}

\subsection{Main Results}
\subsubsection{Overall Performance}

\paragraph{Domain experts excel at their respective tasks.} We first analyze how
expert LLMs specialize to specific domains. Results are summarized in \autoref{table:single_expert}.
As expected, individual expert LLMs achieve the best performance in their respective domain, where the math and code domains see especially large improvements.
In addition, there are several interesting observations.
We see that the math expert training improved its code performance as well, indicating a close relation of these domains.
However, such single-domain continued training also suffers from catastrophic forgetting with significant performance drops on some tasks in other domains. For example, the math and code expert are much worse on TriviaQA than the seed model.

\paragraph{BTX improves all tasks where experts specialize.} \autoref{table:btx_overall} and \autoref{fig:plots} (right) show aggregated performance across multiple domains. More detailed per-task results are reported in \autoref{table:btx_overall_appendix} in the Appendix. 
Compared to the seed model \textsc{Llama-2 7B}, BTX models (both Sample Top-1 and Top-2 corresponding to different number of active parameters) improve on all expert domains, such as math, coding and world knowledge without regressing on other tasks such as commonsense reasoning. 
BTX with Top-2 experts (our default) also approaches the best performance of the specialized models \textsc{Llemma 7B} and  \textsc{CodeLlama 7B} in the math and coding domains, while drastically improving over those models on domains that are not their speciality such as world knowledge and commonsense reasoning. 
Compared to alternative data-matching (DM) methods for continued pretraining such as dense and sparse upcycling, BTX achieves better performance on average with small gaps in the math and coding domains. BTX outperforms BTM by a large margin on average,
indicating that MoE finetuning to learn token-level routing is beneficial.  Overall, the results demonstrate that BTX is a more compute efficient method for continued pretraining which is robust to task interference from multi-task learning. 
BTX also outperforms
\textsc{Llama-2 13B} on all tasks except 
reasoning, even though \textsc{Llama-2 13B}
uses significantly more training compute and has slightly more active parameters.

\begin{table*}[t]
    \centering
    \footnotesize
    \setlength\tabcolsep{2pt}
    \begin{tabular}{lcccc|cccccc}
    \toprule
     & MoE & Training & Total compute & \#tokens & Math & Code & Knowledge & Reasoning & MMLU & Average \\ 
     & compute & time (days) & (GPU-days) & (B) &  \bf  & \bf  & \bf  & \bf  & \bf  & \bf  \\ 
    \midrule
    BTX & \phantom{0}23\% & 7.8 & \phantom{0}926.1 & 533 &  27.4 & 34.0 &  41.0 & 63.5 & 52.5 & 47.9  \\
    Sparse upcycling (CM) & 100\% & 7.9 & 1007.1 & 252 &  28.2 & 30.7 & 41.3 & 62.9 & 52.1 &  47.3  \\
   
    \bottomrule
    \end{tabular}
    \caption{Comparison between BTX and Sparse upcycling with compute-matching (CM), which is a special case of BTX without the expert training stage as is shown by the first column that 100\% of compute is spent on MoE training. We also report total training time,  compute and number of training tokens. Comparing both performance on individual domains as well as the average, we can see that BTX has  more balanced performance, in addition to higher throughput.}
    \label{table:su_compute_matching}
\end{table*}

We further compare BTX with the sparse upcycling baseline in the compute-matching (CM) scenario. Both train on the same data mixture during the MoE stage, but differ in terms of the percent of compute spent on MoE training. 
While sparse cycling performs close behind BTX, the parallel training of experts increases the training throughput of BTX, as is shown in \autoref{table:su_compute_matching}. As a result, BTX can train with more than $2\times$ the data than pure MoE given the same training compute budget, and achieves slightly higher average performance across all domains.

\footnotetext{The GPU days of Llama-2 13B is an approximate measurement, calculated by doubling the training compute of a 7B model trained with the same amount of pretraining data  (according to \citet{touvron2023llama2} Table 2). Since Llama-2 13B is not trained from the seed model, we simply report their difference in GPU days.
}
\begin{table*}[t]
    \centering
    \small
    \begin{tabular}{lcccc}
    \toprule
    \multirow{2}{*}{\bf Routing method} & \multicolumn{2}{c}{\bf Active parameters (B)} & \multirow{2}{8em}{\centering \bf MoE Finetune tokens (B)} & \multirow{2}{4em}{\centering \bf Average score} \\
    \cmidrule(lr){2-3}
     & \bf Training  & \bf Inference &  &    \\    
    \midrule
    \midrule
    Switch Top-1 & \phantom{0}6.7 & \phantom{0}6.7 & \phantom{0}10 & 24.7    \\
    Sample Top-1 & \phantom{0}6.7 & \phantom{0}6.7 & \phantom{0}10 & 33.0  \\
    Top-2 & 11.1 & 11.1 & \phantom{0}10 &  34.6   \\
    Soft routing & 19.7 & 19.7 & \phantom{0}10 & 35.8   \\
    \midrule
    Sample Top-1 & \phantom{0}6.7 & \phantom{0}6.7 & \phantom{0}40 & 35.3  \\
    Top-2 & 11.1 & 11.1 & \phantom{0}40 &  35.9   \\
    Soft routing & 19.7 & 19.7 & \phantom{0}40 & 37.3   \\
    \midrule
    Sample Top-1 & \phantom{0}6.7 & \phantom{0}6.7 & 160 & 36.9  \\
    Top-2 & 11.1 & 11.1 & \phantom{0}80 &  37.3   \\
    \bottomrule
    \end{tabular}
    \caption{Ablations on different routing methods during BTX training. Average score is based on performance on representative tasks including GSM8K, HumanEval, Natural Questions, ARC Challenge and MMLU.}
    \label{table:ablation}
\end{table*}

\begin{table*}[t]
    \centering
    \small
    \begin{tabular}{lcccccc}
    \toprule
     & \bf GSM8K  & \bf Human & \bf Natural & \bf ARC & \bf MMLU & \bf Average \\
     &  & \bf Eval & \bf Questions & \bf Challenge & & \bf Score   \\    
    \midrule
    \midrule
    BTX & 29.8 & 27.4 & 23.0 & 43.4 & 50.0 & 34.7   \\
    \midrule
    no load-balancing (LB) & 34.6 & 19.5 & 23.2 & 44.4 & 51.6 & 34.6   \\
    no LB \& freeze experts & 34.8 & 18.3 & 24.1 & 44.9 & 51.4 & 34.7 \\
    blending experts & 13.9  & 17.1  & 9.9 &  34.1 & 36.2 &  22.2  \\
    split experts, top-2 of 8 & 22.0  & 20.1 & 16.8 & 39.1 & 41.8 & 28.0   \\
    split experts, top-4 of 8 & 29.6  & 26.8 & 22.9 & 44.0 & 49.4 & 34.5   \\
    \bottomrule
    \end{tabular}
    \caption{Ablations on different BTX training strategies. All variants are initialized from the same experts and trained for a total of 10B tokens during MoE finetuning.}
    \label{table:additional_ablation}
\end{table*}

\subsubsection{Better compute-performance tradeoff}
We compare BTX with baselines in terms of compute efficiency in \autoref{fig:plots} (left). The X-axis shows the total training compute starting from the seed model measured in GPU days, which includes the domain expert training and finetuning of the MoE model. The Y-axis measures the overall performance reported in \autoref{table:btx_overall}.

\paragraph{Better performance than dense and BTM.} Despite that the MoE training stage uses a fraction of the total training budget in pretraining (for example, \textsc{Llama-2} pretraining uses 2T tokens), BTX brings steep improvements on general capabilities compared to alternative continued pretraining approaches such as multi-task learning of the dense model and Branch-Train-Merge.

\paragraph{More efficient than sparse upcycling.} As a special case of BTX, sparse upcycling without expert training outperforms dense and BTM but not BTX, given the same or larger compute budget.
The compute efficiency gains of BTX are from the embarrassingly parallel training of experts before MoE finetuning. 

In terms of the active number of parameters (shown as circle sizes in \ref{fig:plots} (left)), the MoE models are similar to the \textsc{Llama-2 13B} model. BTX uses less than half of the additional training compute compared to \textsc{Llama-2 13B}, but demonstrates improved performance on expert domains (math, code, and knowledge) and achieves better overall performance. This indicates that BTX's training is more effective for the late stage of pretraining than using the same training protocol throughout the entire of pretraining.

\subsection{Ablations \& Analysis}
\label{sec:ablations}
\subsubsection{Ablations of BTX training}
\label{sec:ablations_of_btx}
First, we compare the different routing methods with varying amount of active parameters for different amounts of finetuning. For fair comparison, load balancing is not used in any of them.
Results are shown in \autoref{table:ablation}.
For Switch routing, we set its capacity factor to 1.5 (a hard limit after which routed tokens will be dropped). We found the Switch router to be subpar in average performance.
The soft routing performs the best, but that is expected since it lacks sparsity and has the highest number of active parameters.
Overall, the Top-2 routing gives us a good balance between performance and efficiency.

We also ablate additional design choices of BTX, with results summarized in \autoref{table:additional_ablation}. 
We found that MoE training without load balancing performs worse on the coding task (HumanEval), but has higher math (GSM8k) accuracy. 
The routing analysis in the next section will give more insight into this trade-off. 
Next, freezing the feedforward modules initialized from each expert, and only training the rest of the MoE model has little impact on performance across all tasks.
This suggests that individual experts already gained sufficient domain knowledge during the branch-train stage, while the mix (MoE finetuning) stage mainly trains the other parameters such as averaged weights in the self-attention and the router transformations $W_i$.

We also test our blending and splitting techniques described in \Cref{sec:variation}.
The performance across all tasks dropped when experts are mixed, suggesting that domain FF layers cannot be mixed in this way.
Splitting each domain FF into $C=2$ chunks to obtain 8 modules in the MoE layer also does not improve performance, even if Top-4 routing is used to match the active number of parameters.

\begin{figure}[t]
    \centering
    \includegraphics[width=0.95\linewidth]{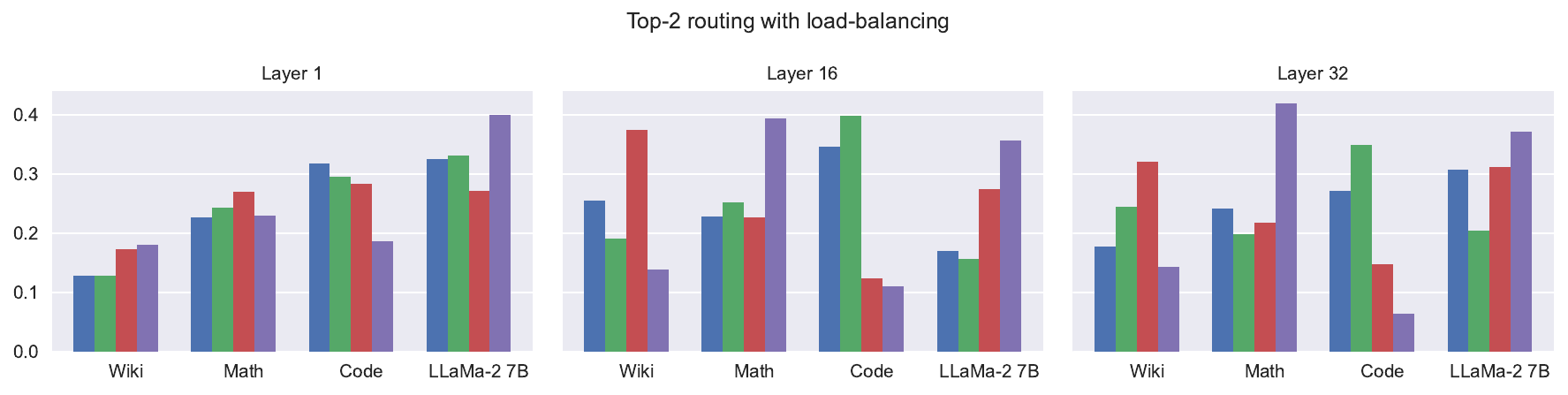}
    \includegraphics[width=0.95\linewidth]{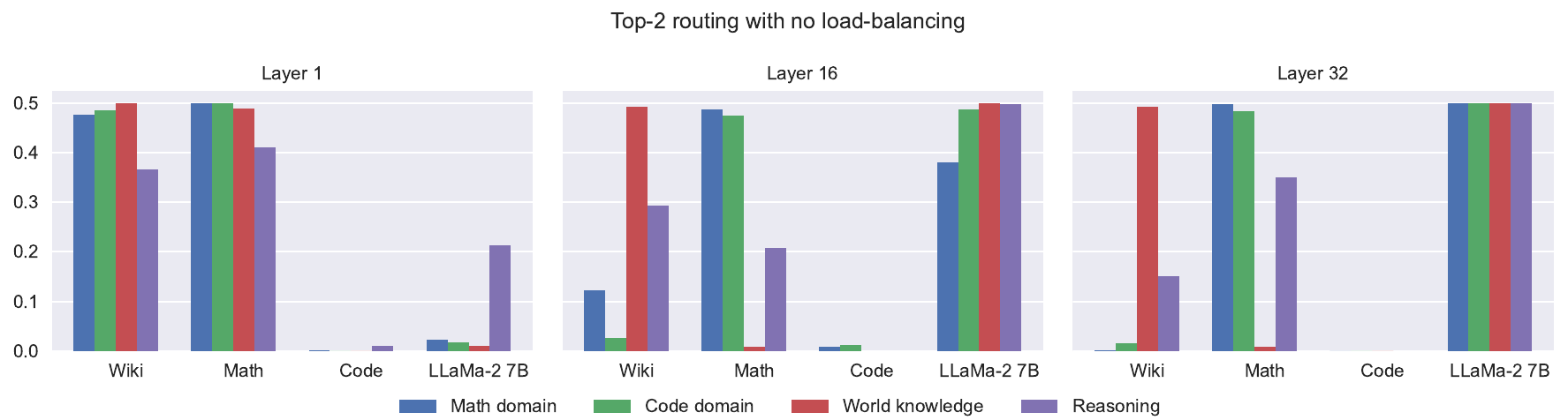}
    \caption{BTX routing decisions of the tokens at various layers to different experts (Wiki, Math, Code, \textsc{LLaMa-2 7B}) for different downstream tasks. The tasks are aggregated by domain: Code (Human Eval, MBPP), Math (GSM8K, MATH), World knowledge (Natural Questions, TriviaQA), and Reasoning (ARC-Easy, ARC-Challenge, SIQA, PIQA, and WinoGrande). We observe that Top-2 routing with load balancing (top) ensures a more uniform distribution of the load between experts compared to Top-2 without load balancing (bottom).}
    \label{fig:routing_3}
\end{figure}

\subsubsection{Routing Analysis}
To gain an in-depth understanding of the performance of BTX, we run model evaluations on downstream tasks and examine the routing decisions among the experts. The results are summarized in \autoref{fig:routing_3}, and we also report detailed ablation results for different BTX setups in \autoref{appendix_routing_analysis}.  Compared to  other routing methods, Top-2 routing with load balancing ensures a more uniform distribution of the load between experts. Analyzing the token probability distributions, we observe a shift towards low probability scores across all experts with load balancing, especially closer to the final layers of the model, which contributes to the fair routing.  Interestingly, all models without load balance heavily rely on the Math expert, with a low overall contribution from other experts, especially the Code expert. A dead Code expert comes ``back to life'' with load balancing introduced in training. In fact, it not only becomes visible, but becomes the dominant expert in the math and code domains.

Examples of the routing decisions for Top-2 with load balancing can be found in the \autoref{table:token_routing}. Overall across math domain tasks, tokens are often routed to the Code and \textsc{Llama-2 7B} experts. If we look at a more detailed token distribution (\autoref{appendix_routing_analysis}, \autoref{fig:routing_t2lb_per_task}), we find that the GSM8K task prefers Code and \textsc{Llama-2} experts, while the MATH task relies more on the in-domain Math expert. We hypothesise that this happens because the GSM8K dataset consists of grade school math problems that require common sense knowledge and basic arithmetic operations.
Both the Code and World knowledge tasks mostly route to the in-domain Code and Wikipedia experts respectively. 
As observed earlier in Section~\ref{sec:ablations_of_btx}, when load balancing is introduced, there are improvements in coding tasks but degradation in math tasks, which can be explained with these changes in domain expert routing. The reasoning tasks in contrast exhibit similar behaviour, and rely equally on Math and generalist LLM’s expertise.

\input{large_tables/table_routing_examples}

\section{Conclusion}
We introduced Branch-Train-MiX (BTX), a simple continued pretraining method to improve an LLM's capabilities. It trains multiple copies of a seed LLM to specialize in multiple domains in an asynchronous and parallel fashion and later merges them back into a single Mixture-of-Experts (MoE) model via finetuning.
While the initial parallel training stage brings higher training throughput and scalability, the second MoE finetuning stage makes the final LLM more performant.
Our experiments suggest that a generalist LLM's performance can be boosted by continued training on datasets with specialized knowledge and skills using our method. We find that the BTX approach is more compute efficient than training a larger generalist LLM or several separately specialized LLMs. These insights can inform how to allocate compute in late pretraining to achieve a strong generalist model.

\section{Limitations \& Future Work} 
Although our experimental results on BTX are promising, we have not fully explored its potential in this paper.
Due to compute limitations, we only experimented with three domains and four experts in this paper.
Training on more domains such as using unsupervised domain discovery \citep{gururangan2023scaling} should amplify the benefit of the parallelization of experts training.
Having more experts will also make the final MoE model more efficient because the number of active experts can remain the same while its overall capacity increases.
In our experiments, we used a simple implementation of MoE and did not optimize it using more complex techniques such as placing different experts on different GPUs to run them in parallel.
Such an efficient MoE implementation could shorten the training time of BTX, and the sparse upcycling baseline as well.

Compared to BTM, BTX provides an approach to finetune the combined experts, which can be directly applied in instruction finetuning or RLHF procedures.
However, we leave that for future work as we focused on the pretraining stage in this paper.

The question of whether experts in MoE are better off specializing in specific domains or not is an interesting one that is worth further investigation.
Our approach explicitly tied experts to certain domains, but such specialization does not seem to emerge naturally during MoE training \citep{Jiang2024MixtralOE}.
We observed that some experts are used more in their corresponding domain tasks, showing that their domain specialization partially remains even after the MoE finetuning. 

We only compared BTX to two of its special variants, i.e. BTM with 100\% compute allocated to expert training and 0\% on MoE finetuning, and sparse upcycling with 0\% compute allocated to expert training and 100\% on MoE finetuning. Future work could perform a thorough sweep of the compute allocation ratio between expert training and MoE training. Also, we did not perform experiments with different data mixtures for MoE finetuning other than uniform sampling.

\section{Acknowledgements}
We thank Margaret Li, Kushal Tirumala, Luke Zettlemoyer, Artidoro Pagnoni, Suchin Gururangan, Mike Lewis and Emily Dinan for their discussion and feedback, and Andrew Cohen and Arun Babu for their help with the training implementation.

\clearpage
\newpage
\bibliographystyle{assets/plainnat}
\bibliography{paper}

\clearpage
\newpage
\beginappendix

\section{Data mixture}
\label{appendix:data}
\autoref{table:data_mixture} shows the exact data mixture ratios used in training each domain expert.
For finetuning the MoE model, we sample datasets that used to train math expert, code expert, wikipedia expert and the original \textsc{Llama-2 7B} with probabilities 30.16\%, 40.31\%, 10.30\% and 19.23\%. 

\begin{table*}[h]
    \centering
    \small
    \begin{tabular}{llc}
    \toprule
    Domain & Dataset & Sampling ratio (\%)   \\ 
    \midrule
    \multirow{5}{*}{Math}
    & AlgebraicStack & 13.57 \\
    & OpenWebMath & 54.27 \\
    & Arxiv & 27.14 \\
    & Github & 2.99 \\
    & Commoncrawl & 5.01 \\
    \midrule
    \multirow{3}{*}{Code}
    & Code & 82.18 \\
    & Natural language related to code & 9.90 \\
    & Natural language & 6.93 \\
    \midrule
    \multirow{2}{*}{Wikipedia}
    & Wikipedia & 90.91 \\
    & Commoncrawl & 9.09 \\

    \bottomrule
    \end{tabular}
    \caption{Data sources and weights for domain experts.}
    \label{table:data_mixture}
\end{table*}

\section{Evaluation}
\label{appendix:eval_metrics}
We use the same evaluation metrics as is used in \cite{touvron2023llama2} and \cite{Rozire2023CodeLO}: for code tasks (HumanEval and MBPP) we report pass@1, for math tasks (GSM8k and MATH) and knowledge tasks (Natural Questions and TriviaQA) we report exact match, we report accuracy for MMLU and ARC. We use greedy decoding for all generations. Detailed results on all tasks are reported in \autoref{table:btx_overall_appendix}.
\begin{table*}[h]
    \centering
    \footnotesize
    \addtolength{\tabcolsep}{-0.4em}
    \begin{tabular}{lcccccccccccc}
    \toprule
    & GSM8K & MATH & Human & MBPP & Natural             & Trivia & ARC-e & ARC-c & Wino & SIQA & PIQA & MMLU  \\
    &       &      & Eval  &     & Questions & QA            &    &      \\    
    \midrule
    \textit{Specialized LLMs} &  &  &  &  &  &  & &  &  \\   
    \textsc{CodeLlama 7B} & 13.0 & 3.3 & 31.1 & 41.4 & 11.5 & 32.8 & 67.4 & 34.0 & 62.7	& 46.1 & 72.9 & 38.6  \\ 
    \textsc{Llemma 7B} & 39.3 & 16.7 & 25.6 & 41.4 & 9.4 & 24.9 & 28.7 & 26.8 & 50.1 & 37.3	& 51.0 & 33.5 \\ 
     \midrule
    \textit{Generalist LLMs} &  &  &  &  &  &  & &  &  \\
    \textsc{Llama-2 7B} & 14.7 & 2.5 & 12.8 & 20.8 & 16.4 & 58.5 & 76.4 & 43.8 & 69.2	& 48.3 &	78.8 & 46.1 \\
    \textsc{Llama-2 13B} & 28.7 & 3.9 & 18.3 & 30.6 & 16.1 & 63.8	& 77.3 &  49.4 & 73.0 &	 50.1 & 80.8 & 52.8  \\
    Dense (DM) & 26.7 & 9.9 & 20.7 & 30.8 & 24.0 & 55.3 & 76.7 & 44.5 & 68.9 &	48.3 &	78.2 & 49.8  \\
    Sparse upcycling (DM), Top-2 & 37.3 & 18.9 & 29.3 & 40.2 & 18.8 & 49.2 &	76.3 & 43.4 & 66.4 &	47.3	& 77.9 & 51.1  \\ 
    Sparse upcycling (CM), Top-2 & 40.1	& 16.2	& 26.2 &	35.2 & 24.5 & 58.2 & 75.6 & 44.7 & 69.1	& 47.1 & 78.0 & 52.1  \\
    BTM, Top-1 & 27.4 & 15.2 &  30.8 & 41.9 & 15.0 & 38.0 & 72.8 &  38.1 & 68.4& 47.8 & 77.9 & 44.3  \\
    BTM, Top-2 & 27.7 & 15.3 & 30.6 & 42.6 & 15.3 & 38.5  & 73.1 & 38.5 & 68.3& 48.0 & 78.1 & 44.3  \\
    BTX, sample Top-1 & 36.9 & 15.8 & 25.6 & 37.4 & 23.7 & 56.4	& 76.7 & 45.0 & 70.6 &	48.0 & 78.2 &  53.2  \\
    BTX, Top-2 & 37.1 & 17.8 & 28.7 & 39.4 &  24.8 & 57.1	& 76.9 & 45.6 & 67.9 &	48.7	& 78.7 & 52.5  \\
    \bottomrule
    \end{tabular}
    \caption{Individual task performance of BTX and baselines.
    }
    \label{table:btx_overall_appendix}
\end{table*}

\section{Routing analysis}
\label{appendix_routing_analysis}

\begin{figure}[t!]
    \centering
    \includegraphics[width=0.95\linewidth]{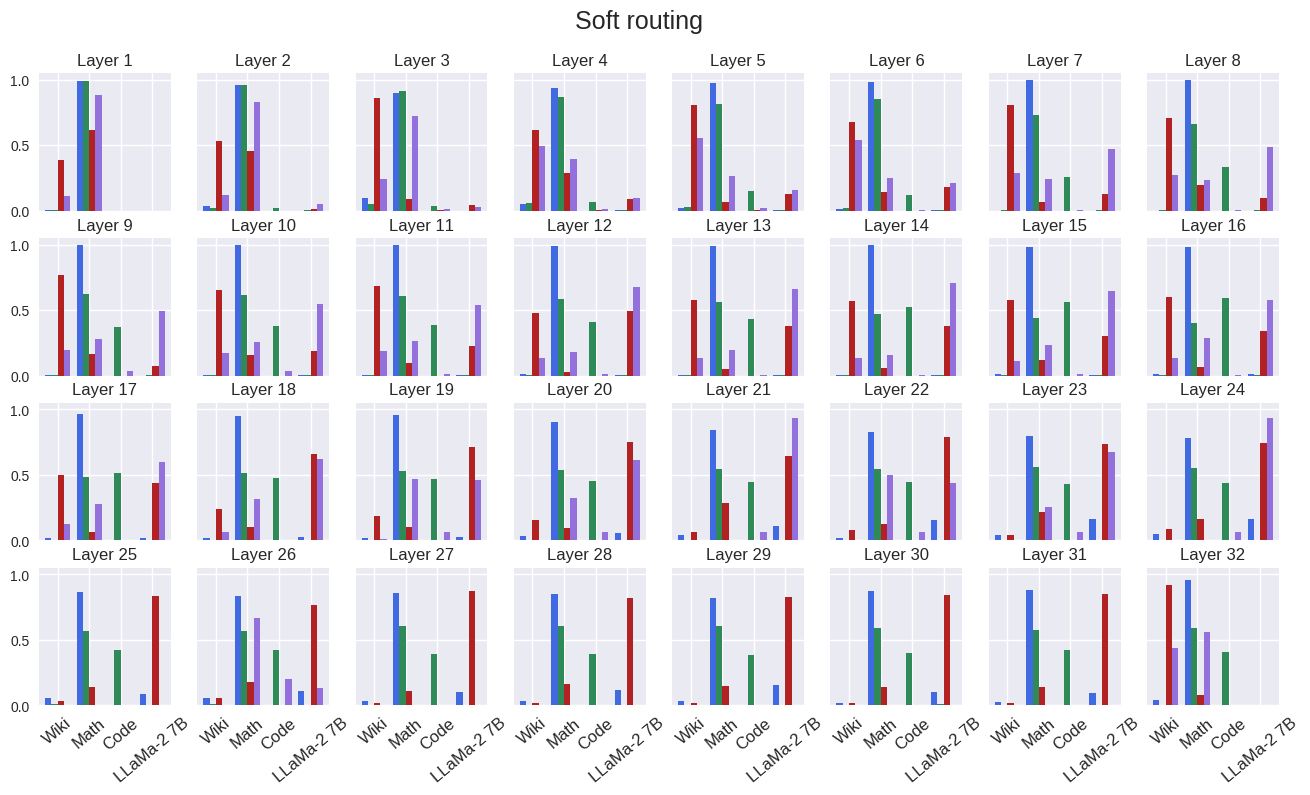}
    \includegraphics[width=0.95\linewidth]{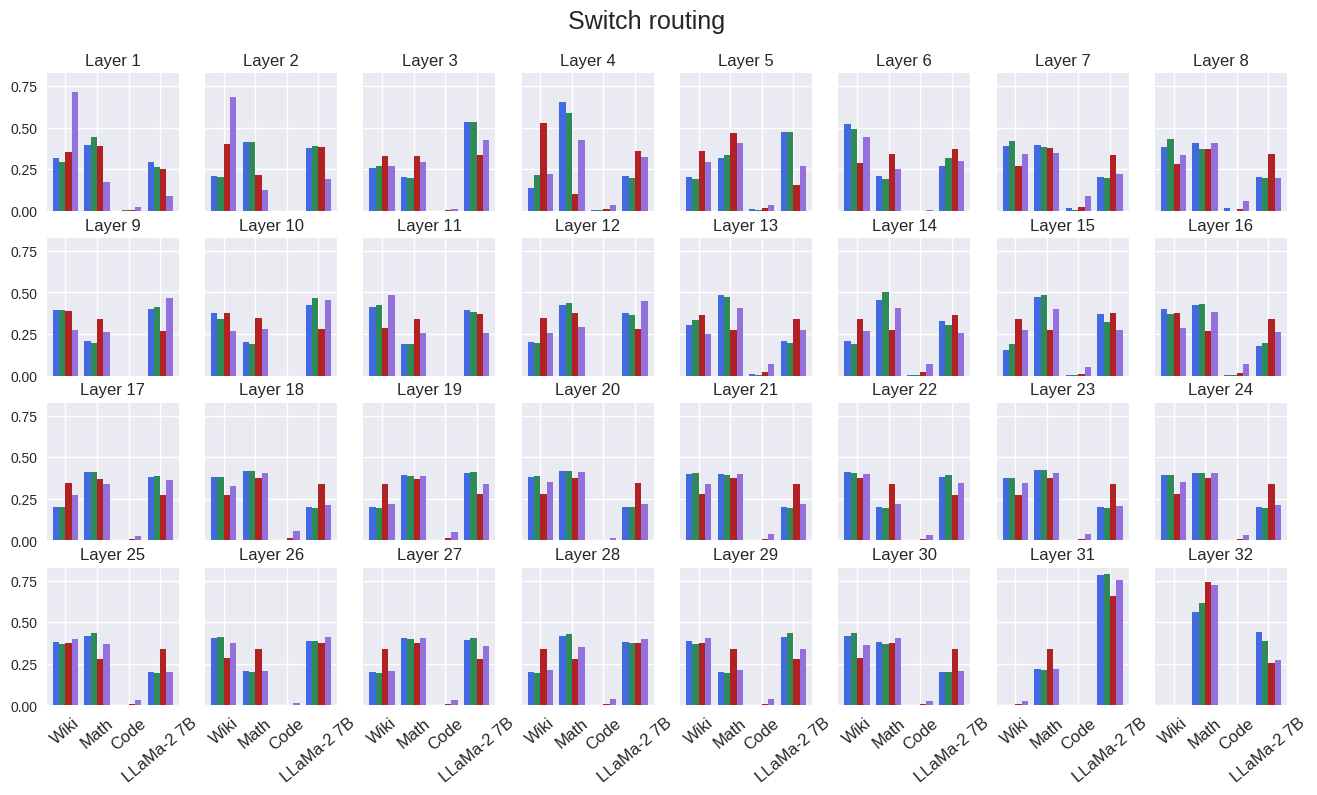}
    \caption{BTX routing decisions of the tokens at various layers to different experts (Wiki, Math, Code, \textsc{LLaMa-2 7B}) for different downstream tasks. The tasks are aggregated by domain: Code (Human Eval, MBPP), Math (GSM8K, MATH), World knowledge (Natural Questions, TriviaQA), and Reasoning (ARC-Easy, ARC-Challenge, SIQA, PIQA, and WinoGrande). We observe that top-2 routing with load balancing ensures more uniform distribution of the load between experts compared to the other routing methods across all layers.}
    \label{fig:routing_all}
\end{figure}

Layer-by-layer comparison of the routing decision  for different router designs and downstream tasks aggregated by task domain is shown in Figure~\ref{fig:routing_all}. Routing distributions slightly vary in the first few layers, but quickly become indistinguishable from layer to layer. One exception is in Switch routing where Math expert becomes dominant across tasks in the last model layer.

\begin{figure}[h]
    \centering
    \includegraphics[width=0.8\linewidth]{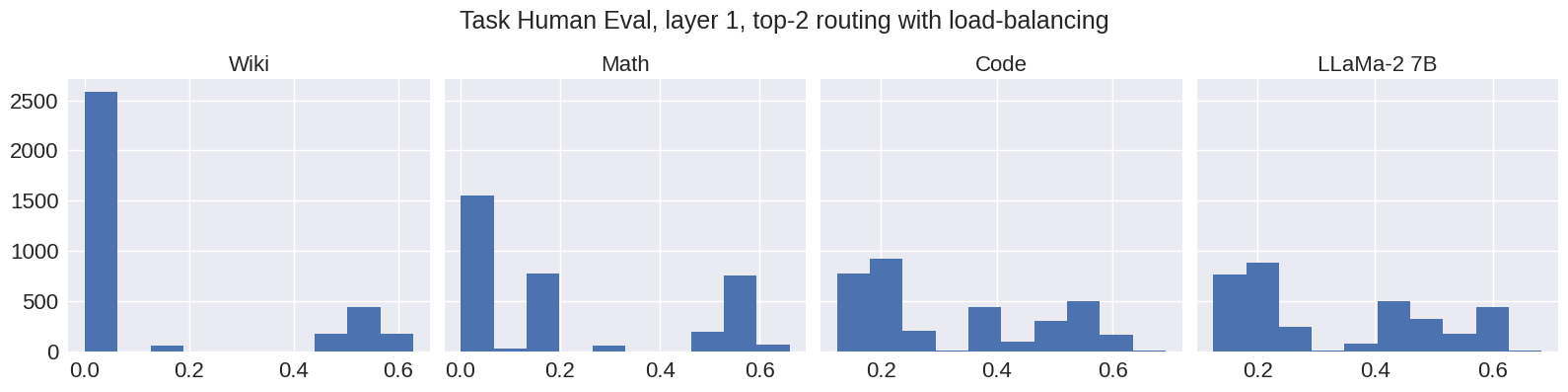}
    \includegraphics[width=0.8\linewidth]{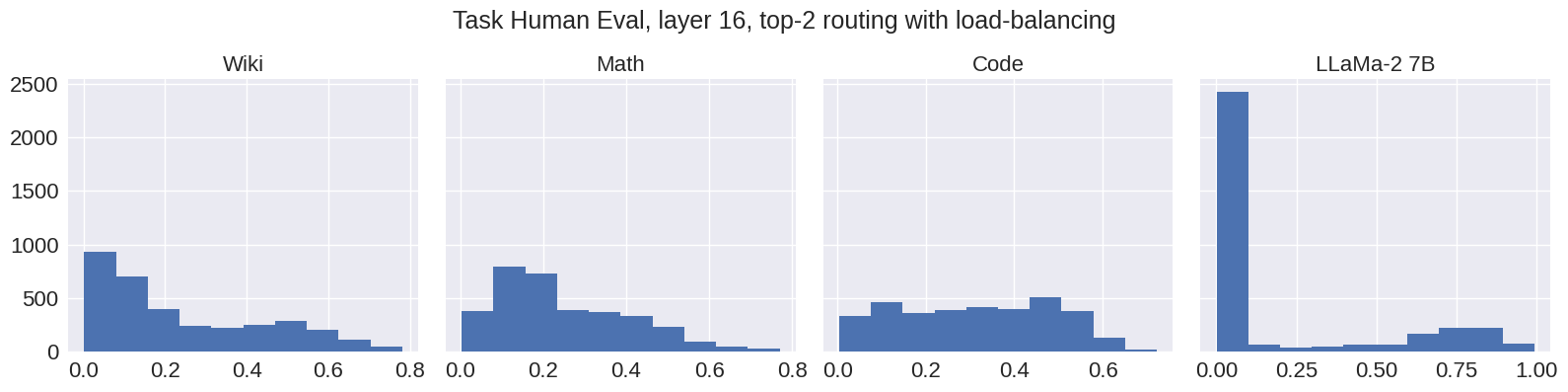}
    \includegraphics[width=0.8\linewidth]{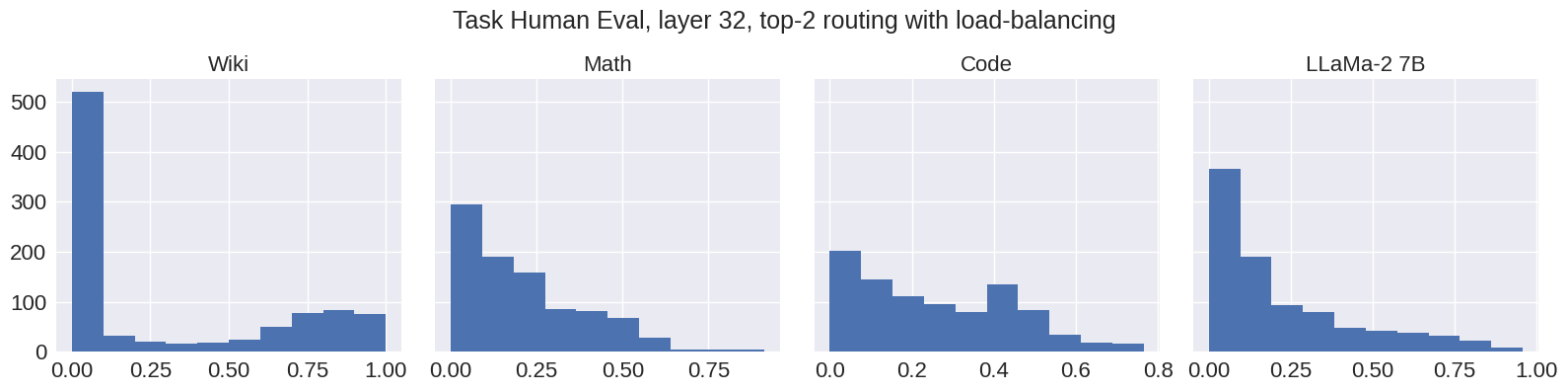}
    
    \includegraphics[width=0.8\linewidth]{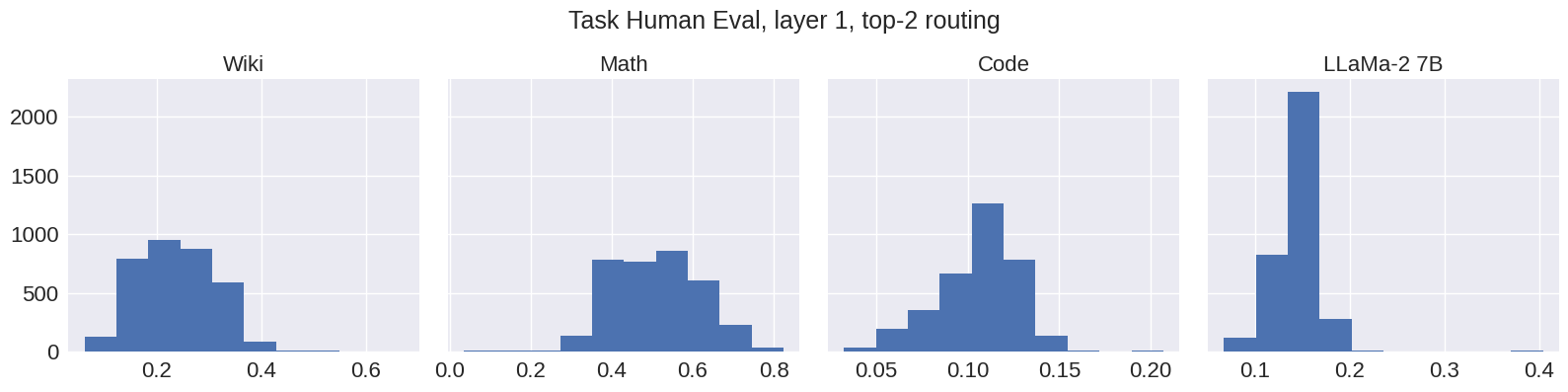}
    \includegraphics[width=0.8\linewidth]{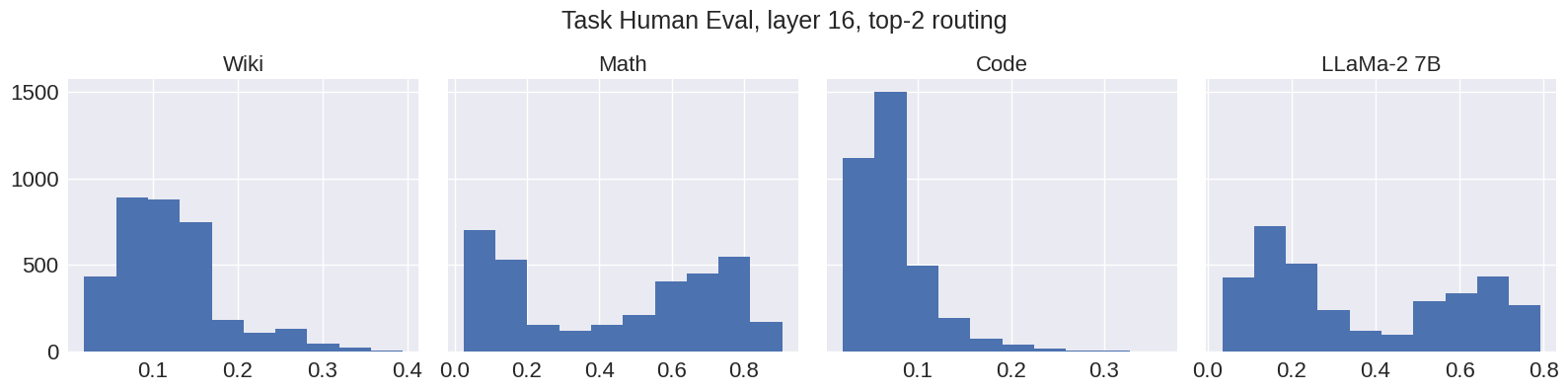}
    \includegraphics[width=0.8\linewidth]{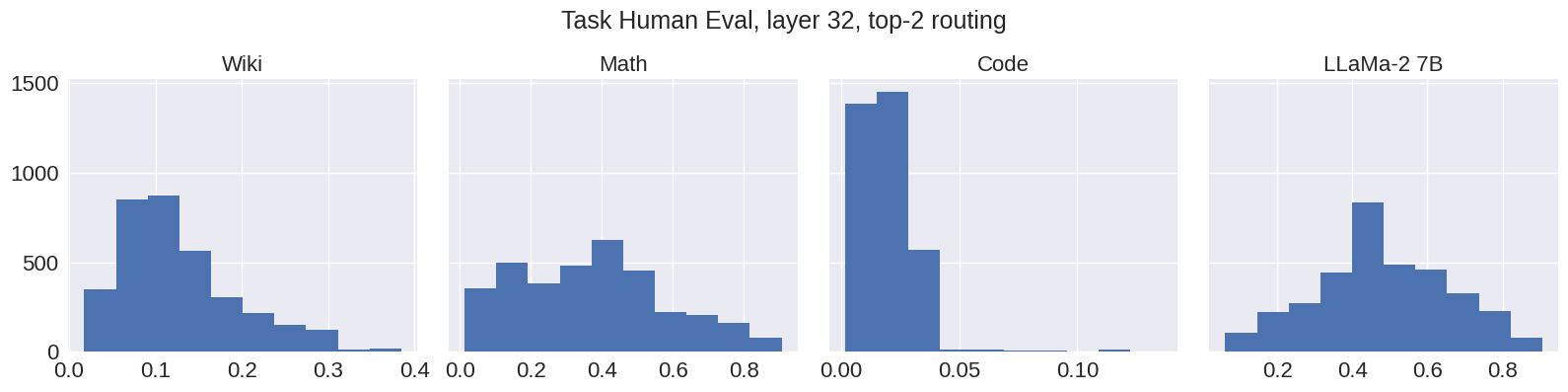}
    \caption{Routing probabilities per expert across different layers for Human Eval task. We compare top-2 routing with (left) and without load balancing (right).}
    \label{fig:hist}
\end{figure}

We observe that Code expert is a dominant force in Code  domain in Top-2 routing with load balancing. Note the difference with other models where load balancing is not added, and Math expert prevails across domains. We look at Code domain closer and compare routing probability distribution for models with and without load balancing in Figure~\ref{fig:hist}. On the bottom three graphs of the picture we can observe a phenomena of the dead expert, where routing probability to Code expert shifted to $0$, while with load balancing added, probability distributions across experts look more similar, with slightly higher expectations for the Code expert.

\begin{figure}[h]
    \centering
    \includegraphics[width=0.8\linewidth]{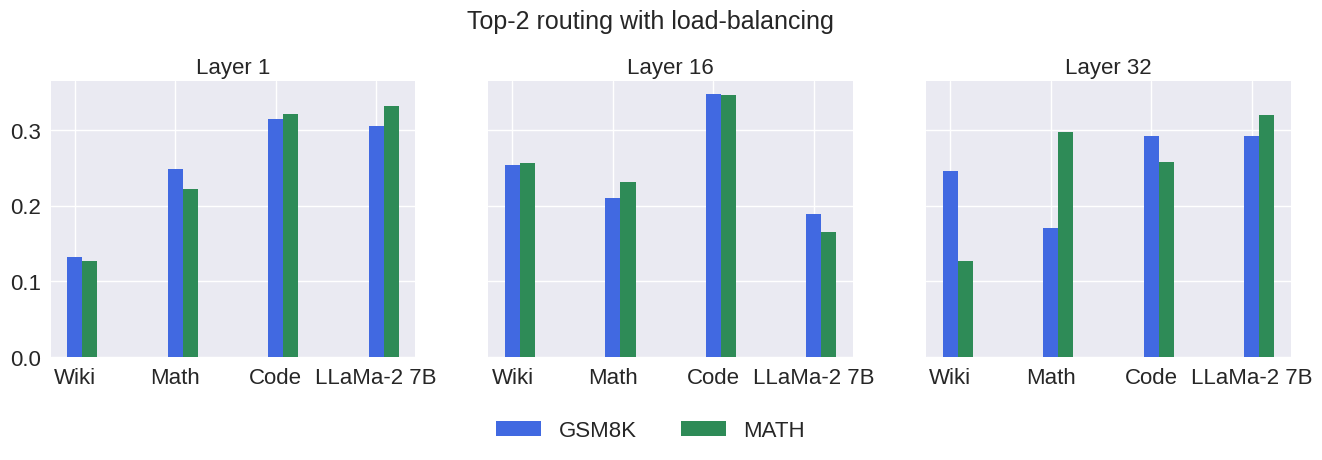}
    \includegraphics[width=0.8\linewidth]{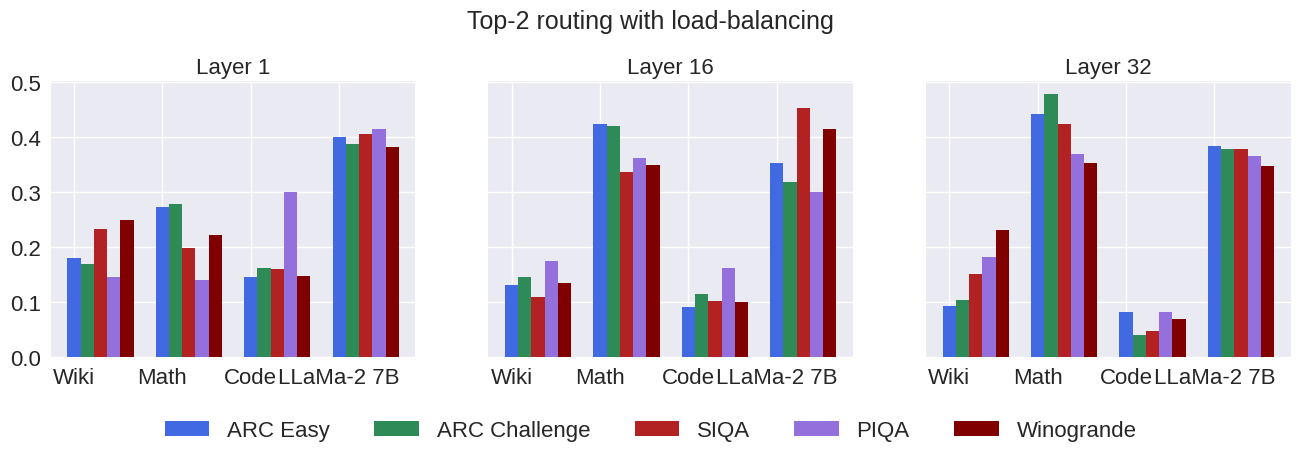}
    \caption{Routing decision of the tokens in Math and Reasoning domains. We observe that GSM8K task prefers Code and \textsc{Llama-2} experts, while MATH task relies more on in-domain expert. In the Reasoning domain, the load is distributed between Math and \textsc{LLaMa-2 7B} experts.}
    \label{fig:routing_t2lb_per_task}
\end{figure}

To understand if experts specialize in other domains, we look closer at per-task distribution. Routing decision of the tokens in Math and Reasoning domains are shown in Figure~\ref{fig:routing_t2lb_per_task}. We observe that GSM8K task prefers Code and \textsc{Llama-2} experts, while Math task more relies on in-domain expert. We hypothesise that this happens because GSM8K dataset consists of grade school math word problems that require common sense knowledge and basic arithmetic operations, while Math task requires college-level math knowledge, and more aligned with Math expert's training data. In the Reasoning domain, all tasks exhibit similar behaviour and equally rely on Math
and generalist LLM’s expertise.

\begin{figure}[h]
    \centering
    \includegraphics[width=0.95\linewidth]{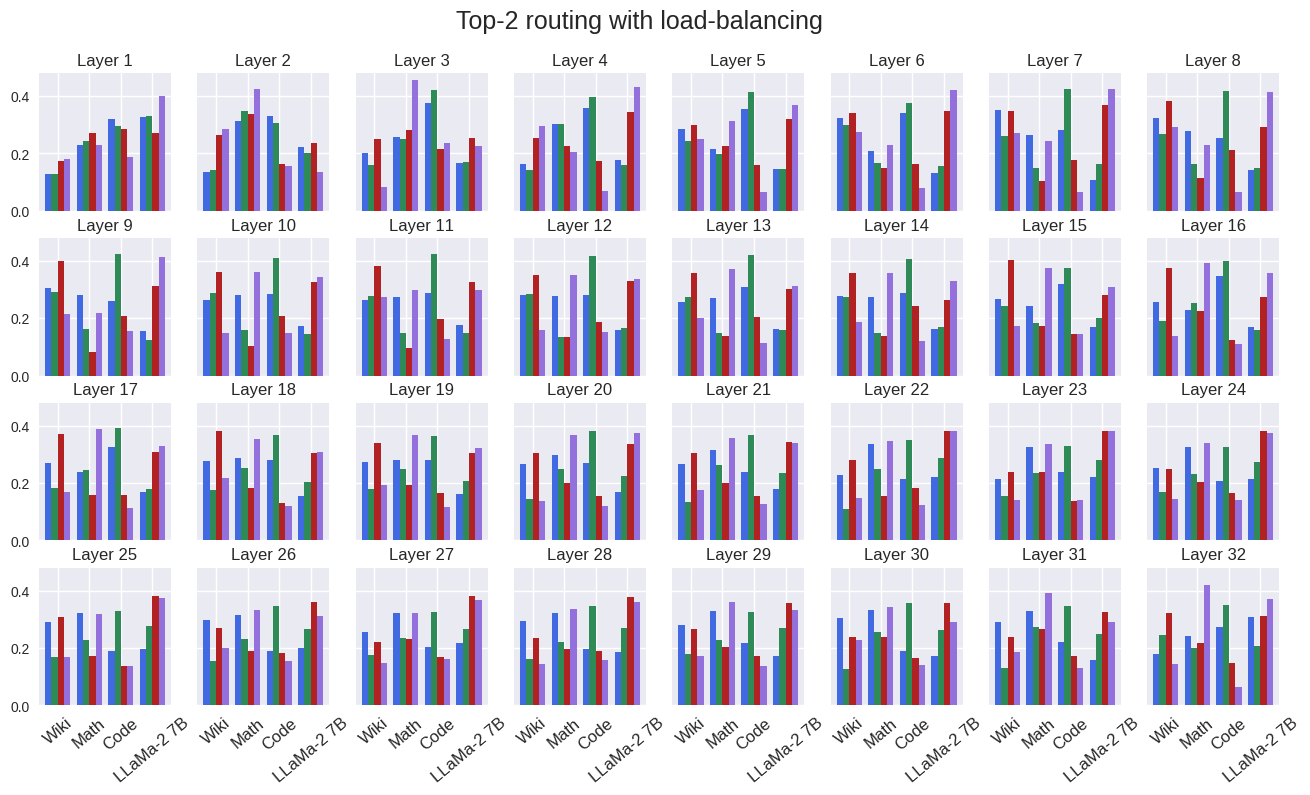} 
    \includegraphics[width=0.95\linewidth]{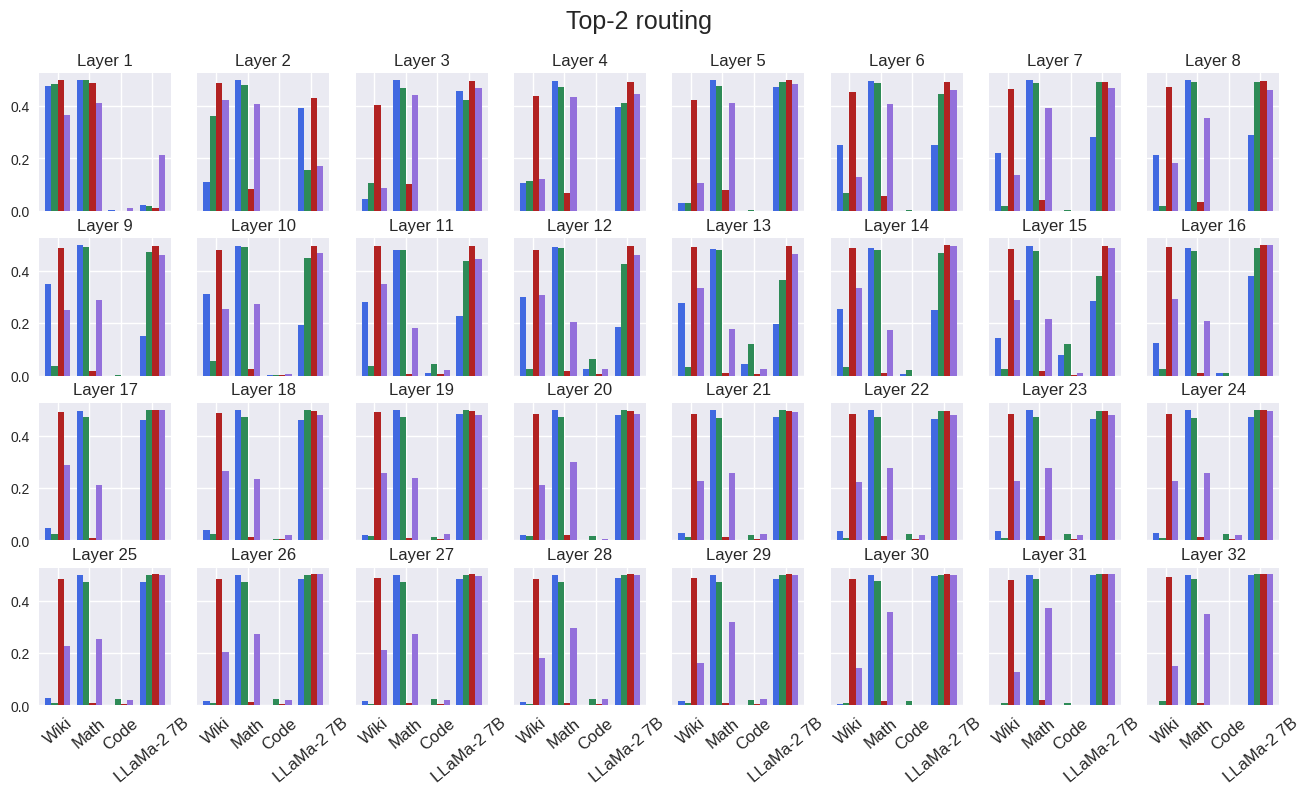} 
\end{figure}

\end{document}

%% file: large_tables/table_routing_examples.tex
\begin{table*}[t]
    \centering
    \scalebox{0.8}{
    \begin{tabular}{ll}
    \toprule
    Task & Question and generation \\   
    \midrule
    GSM8K & \makecell[l]{Q: \underline{\color{blue}{Jan}\color{magenta}{et}\color{blue}{’}}\color{orange}{s} \color{black}{}\underline{\color{magenta}{ducks lay}} \color{orange}{1}\color{black}{}\underline{\color{magenta}{6 eggs} \color{blue}{per} \color{magenta}{day}}\color{cyan}{.} \color{black}{}\underline{\color{blue}{She e}}\color{orange}{ats} \color{black}{}\underline{\color{magenta}{three for breakfast}} \color{cyan}{every} \color{black}{}\underline{\color{magenta}{morning}} \color{cyan}{and} \color{black}{}\underline{\color{magenta}{bakes muff}}\color{orange}{ins} \color{black}{}\underline{\color{blue}{for}}\\ \color{black}{}\underline{\color{blue}{her} \color{magenta}{friends} \color{blue}{every} \color{magenta}{day}} \color{cyan}{with} \color{black}{}\underline{\color{magenta}{four.} \color{blue}{She s}\color{magenta}{ells the remainder} \color{blue}{at} \color{magenta}{the farmers' market}} \color{cyan}{daily for} \color{black}{}\underline{\color{magenta}{\$}}\color{orange}{2} \color{black}{}\underline{\color{magenta}{per fresh}}\\ \color{black}{}\underline{\color{magenta}{du}}\color{orange}{ck} \color{black}{}\underline{\color{magenta}{egg}}\color{red}{.} \color{black}{}\underline{\color{cyan}{How}} \color{orange}{much} \color{black}{}\underline{\color{magenta}{in dollars} \color{blue}{does she} \color{teal}{make} \color{blue}{every}} \color{orange}{day} \color{black}{}\underline{\color{blue}{at the} \color{magenta}{far}}\color{orange}{mers' market}\color{cyan}{?}\\  
    A: \underline{\color{blue}{Jan}}\color{orange}{et}\color{black}{}\underline{\color{magenta}{’}}\color{orange}{s} \color{black}{}\underline{\color{magenta}{du}}\color{orange}{cks} \color{black}{}\underline{\color{magenta}{lay}} \color{orange}{16 eggs} \color{red}{per} \color{orange}{day}. \color{black}{}\underline{\color{blue}{She e}}\color{orange}{ats} \color{black}{}\underline{\color{magenta}{three}} \color{red}{for} \color{orange}{breakfast} \color{red}{every} \color{orange}{morning}\color{cyan}{.} \color{black}{}\underline{\color{blue}{So she has}} \color{orange}{1}\color{black}{}\underline{\color{magenta}{6} \color{teal}{-}} \color{orange}{3} \color{black}{}\underline{\color{magenta}{=}}\\ \color{orange}{13} \color{black}{}\underline{\color{magenta}{eggs} \color{blue}{left. She} \color{magenta}{b}}\color{orange}{akes} \color{red}{m}\color{orange}{uffins} \color{black}{}\underline{\color{blue}{for}} \color{red}{her} \color{black}{}\underline{\color{magenta}{friends}} \color{red}{every} \color{orange}{day} \color{teal}{with} \color{orange}{4}\color{black}{}\underline{\color{magenta}{.}} \color{red}{So she} \color{teal}{has} \color{orange}{13} \color{black}{}\underline{\color{teal}{-}} \color{orange}{4 = 9} \color{black}{}\underline{\color{magenta}{eggs}} \color{red}{left.} \color{black}{}\underline{\color{blue}{She}}\\ \color{black}{}\underline{\color{teal}{s}}\color{orange}{ells} \color{black}{}\underline{\color{blue}{the}} \color{orange}{remainder} \color{red}{at the} \color{black}{}\underline{\color{magenta}{far}}\color{orange}{mers'} \color{black}{}\underline{\color{teal}{market}} \color{red}{daily} \color{cyan}{for} \color{black}{}\underline{\color{teal}{\$}}\color{orange}{2 per fresh duck egg}\color{red}{.} \color{black}{}\underline{\color{blue}{So she has}} \color{orange}{9} \color{black}{}\underline{\color{teal}{*}} \color{orange}{2} \color{black}{}\underline{\color{magenta}{=}} \color{orange}{18} \color{black}{}\underline{\color{magenta}{dollars}\color{blue}{.}}\\ \color{black}{}\underline{\color{blue}{The}} \color{orange}{answer} \color{cyan}{is} \color{orange}{18.}} \\
    \hline
    Human Eval & \makecell[l]{Q: \color{red}{from typing import} \color{blue}{List}\\ \\ \color{black}{}\underline{\color{teal}{def has\_close}\color{orange}{\_}\color{teal}{elements(numbers: List[float}\color{orange}{],} \color{teal}{threshold:} \color{cyan}{float}\color{orange}{)}\color{teal}{-> bool}\color{orange}{:}}\\    
    \quad \color{black}{}\underline{\color{teal}{""" Check if in given list}} \color{blue}{of} \color{black}{}\underline{\color{teal}{numbers}}\color{blue}{,} \color{black}{}\underline{\color{teal}{are any}} \color{magenta}{two} \color{black}{}\underline{\color{teal}{numbers closer}} \color{blue}{to} \color{magenta}{each} \color{red}{other} \color{blue}{than} \\
    \quad \color{black}{}\underline{\color{teal}{given threshold.}}\\ 
    \quad \color{black}{}\underline{\color{teal}{> > > has}}\color{red}{\_close\_}\color{blue}{elements}\color{black}{}\underline{\color{teal}{([1.0}\color{orange}{,} \color{teal}{2}\color{orange}{.0, 3.0],} \color{teal}{0}\color{orange}{.}\color{teal}{5)}}\\
    \quad \underline{\color{teal}{False}}\\
    \quad \underline{\color{teal}{> > > has}\color{cyan}{\_close}}\color{red}{\_}\color{black}{}\underline{\color{cyan}{elements}\color{orange}{([}\color{teal}{1}\color{orange}{.0,} \color{teal}{2}\color{orange}{.8}\color{teal}{, 3}\color{orange}{.}\color{teal}{0}\color{orange}{,} \color{teal}{4}\color{orange}{.}\color{teal}{0,} \color{orange}{5.0}\color{teal}{,} \color{orange}{2}\color{teal}{.}\color{orange}{0}\color{teal}{],}} \color{red}{0}\color{black}{}\underline{\color{teal}{.}\color{orange}{3)}}\\
    \quad \underline{\color{orange}{True}}\\
    \quad \underline{\color{orange}{"""}}\\
    A: \underline{\color{teal}{for} \color{teal}{i} \color{teal}{in}} \color{red}{range}\color{black}{}\underline{\color{orange}{(}\color{teal}{len}\color{orange}{(}}\color{red}{numbers}\color{black}{}\underline{\color{teal}{)} \color{teal}{-} \color{orange}{1}\color{orange}{):}}\\
    \quad \underline{\color{teal}{for} \color{teal}{j} \color{orange}{in} \color{blue}{range}\color{orange}{(}\color{teal}{i} \color{teal}{+}} \color{red}{1}\color{black}{}\underline{\color{orange}{,} \color{teal}{len}\color{orange}{(}}\color{blue}{numbers}\color{black}{}\underline{\color{teal}{)):}}\\
    \quad \quad \underline{\color{teal}{if} \color{teal}{abs}\color{teal}{(}\color{teal}{numbers}\color{orange}{[}\color{orange}{i}\color{orange}{]} \color{orange}{-} \color{orange}{numbers}\color{cyan}{[j]}\color{orange}{)} \color{teal}{< threshold}\color{orange}{:}}\\
    \quad \quad \quad \underline{\color{teal}{return} \color{cyan}{True}}\\
    \underline{\color{teal}{return} \color{cyan}{False}}}\\
    \hline
    \makecell[l]{Natural\\ Questions} & \makecell[l]{Q: \underline{\color{cyan}{who} \color{blue}{got} \color{red}{the first no}}\color{magenta}{bel prize} \color{black}{}\underline{\color{blue}{in}} \color{magenta}{physics}\\ 
    A: \underline{\color{red}{Max}} \color{magenta}{Plan}\color{orange}{ck}}\\
    \bottomrule
    \end{tabular}}
    \caption{Examples of the token routing decisions for the Top-2 routing with load balancing in the math (GSM8K), code (Human Eval), and knowledge (Natural Questions) domains. Tokens highlighted are routed to the following experts: \color{red}{Wikipedia and \textsc{LLaMa-2 7B}}, \color{magenta}{Math and \textsc{LLaMa-2 7B}}, \color{orange}{Code and \textsc{LLaMa-2 7B}}, \color{teal}{Math and Code}, \color{blue}{Wikipedia and Math}, \color{cyan}{Wikipedia and Code}. \color{black}{Tokens that were routed to the in-domain expert are underlined.}}
    \label{table:token_routing}
\end{table*}

%% file: paper.bbl
\begin{thebibliography}{42}
\providecommand{\natexlab}[1]{#1}
\providecommand{\url}[1]{\texttt{#1}}
\expandafter\ifx\csname urlstyle\endcsname\relax
  \providecommand{\doi}[1]{doi: #1}\else
  \providecommand{\doi}{doi: \begingroup \urlstyle{rm}\Url}\fi

\bibitem[Achiam et~al.(2023)Achiam, Adler, Agarwal, Ahmad, Akkaya, Aleman, Almeida, Altenschmidt, Altman, Anadkat, et~al.]{achiam2023gpt}
Josh Achiam, Steven Adler, Sandhini Agarwal, Lama Ahmad, Ilge Akkaya, Florencia~Leoni Aleman, Diogo Almeida, Janko Altenschmidt, Sam Altman, Shyamal Anadkat, et~al.
\newblock Gpt-4 technical report.
\newblock \emph{arXiv preprint arXiv:2303.08774}, 2023.

\bibitem[Aljundi et~al.(2016)Aljundi, Chakravarty, and Tuytelaars]{Aljundi2016ExpertGL}
Rahaf Aljundi, Punarjay Chakravarty, and Tinne Tuytelaars.
\newblock Expert gate: Lifelong learning with a network of experts.
\newblock \emph{2017 IEEE Conference on Computer Vision and Pattern Recognition (CVPR)}, pages 7120--7129, 2016.
\newblock \url{https://api.semanticscholar.org/CorpusID:914027}.

\bibitem[Austin et~al.(2021)Austin, Odena, Nye, Bosma, Michalewski, Dohan, Jiang, Cai, Terry, Le, and Sutton]{Austin2021ProgramSW}
Jacob Austin, Augustus Odena, Maxwell Nye, Maarten Bosma, Henryk Michalewski, David Dohan, Ellen Jiang, Carrie~J. Cai, Michael Terry, Quoc~V. Le, and Charles Sutton.
\newblock Program synthesis with large language models.
\newblock \emph{ArXiv}, abs/2108.07732, 2021.
\newblock \url{https://api.semanticscholar.org/CorpusID:237142385}.

\bibitem[Awasthi and Sarawagi(2019)]{awasthi2019continual}
Abhijeet Awasthi and Sunita Sarawagi.
\newblock Continual learning with neural networks: A review.
\newblock In \emph{Proceedings of the ACM India Joint International Conference on Data Science and Management of Data}, pages 362--365, 2019.

\bibitem[Azerbayev et~al.(2023)Azerbayev, Schoelkopf, Paster, Santos, McAleer, Jiang, Deng, Biderman, and Welleck]{Azerbayev2023LlemmaAO}
Zhangir Azerbayev, Hailey Schoelkopf, Keiran Paster, Marco~Dos Santos, Stephen McAleer, Albert~Q. Jiang, Jia Deng, Stella Biderman, and Sean Welleck.
\newblock Llemma: An open language model for mathematics.
\newblock \emph{ArXiv}, abs/2310.10631, 2023.
\newblock \url{https://api.semanticscholar.org/CorpusID:264172303}.

\bibitem[Bisk et~al.(2020)Bisk, Zellers, Gao, Choi, et~al.]{bisk2020piqa}
Yonatan Bisk, Rowan Zellers, Jianfeng Gao, Yejin Choi, et~al.
\newblock Piqa: Reasoning about physical commonsense in natural language.
\newblock In \emph{Proceedings of the AAAI conference on artificial intelligence}, volume~34, pages 7432--7439, 2020.

\bibitem[Brown et~al.(2020)Brown, Mann, Ryder, Subbiah, Kaplan, Dhariwal, Neelakantan, Shyam, Sastry, Askell, Agarwal, Herbert-Voss, Krueger, Henighan, Child, Ramesh, Ziegler, Wu, Winter, Hesse, Chen, Sigler, Litwin, Gray, Chess, Clark, Berner, McCandlish, Radford, Sutskever, and Amodei]{Brown2020LanguageMA}
Tom~B. Brown, Benjamin Mann, Nick Ryder, Melanie Subbiah, Jared Kaplan, Prafulla Dhariwal, Arvind Neelakantan, Pranav Shyam, Girish Sastry, Amanda Askell, Sandhini Agarwal, Ariel Herbert-Voss, Gretchen Krueger, T.~J. Henighan, Rewon Child, Aditya Ramesh, Daniel~M. Ziegler, Jeff Wu, Clemens Winter, Christopher Hesse, Mark Chen, Eric Sigler, Mateusz Litwin, Scott Gray, Benjamin Chess, Jack Clark, Christopher Berner, Sam McCandlish, Alec Radford, Ilya Sutskever, and Dario Amodei.
\newblock Language models are few-shot learners.
\newblock \emph{ArXiv}, abs/2005.14165, 2020.
\newblock \url{https://api.semanticscholar.org/CorpusID:218971783}.

\bibitem[Chen et~al.(2021)Chen, Tworek, Jun, Yuan, Ponde, Kaplan, Edwards, Burda, Joseph, Brockman, Ray, Puri, Krueger, Petrov, Khlaaf, Sastry, Mishkin, Chan, Gray, Ryder, Pavlov, Power, Kaiser, Bavarian, Winter, Tillet, Such, Cummings, Plappert, Chantzis, Barnes, Herbert-Voss, Guss, Nichol, Babuschkin, Balaji, Jain, Carr, Leike, Achiam, Misra, Morikawa, Radford, Knight, Brundage, Murati, Mayer, Welinder, McGrew, Amodei, McCandlish, Sutskever, and Zaremba]{Chen2021EvaluatingLL}
Mark Chen, Jerry Tworek, Heewoo Jun, Qiming Yuan, Henrique Ponde, Jared Kaplan, Harrison Edwards, Yura Burda, Nicholas Joseph, Greg Brockman, Alex Ray, Raul Puri, Gretchen Krueger, Michael Petrov, Heidy Khlaaf, Girish Sastry, Pamela Mishkin, Brooke Chan, Scott Gray, Nick Ryder, Mikhail Pavlov, Alethea Power, Lukasz Kaiser, Mohammad Bavarian, Clemens Winter, Philippe Tillet, Felipe~Petroski Such, David~W. Cummings, Matthias Plappert, Fotios Chantzis, Elizabeth Barnes, Ariel Herbert-Voss, William~H. Guss, Alex Nichol, Igor Babuschkin, Suchir Balaji, Shantanu Jain, Andrew Carr, Jan Leike, Joshua Achiam, Vedant Misra, Evan Morikawa, Alec Radford, Matthew~M. Knight, Miles Brundage, Mira Murati, Katie Mayer, Peter Welinder, Bob McGrew, Dario Amodei, Sam McCandlish, Ilya Sutskever, and Wojciech Zaremba.
\newblock Evaluating large language models trained on code.
\newblock \emph{ArXiv}, abs/2107.03374, 2021.
\newblock \url{https://api.semanticscholar.org/CorpusID:235755472}.

\bibitem[Clark et~al.(2018)Clark, Cowhey, Etzioni, Khot, Sabharwal, Schoenick, and Tafjord]{Clark2018ThinkYH}
Peter Clark, Isaac Cowhey, Oren Etzioni, Tushar Khot, Ashish Sabharwal, Carissa Schoenick, and Oyvind Tafjord.
\newblock Think you have solved question answering? {T}ry {ARC}, the {AI2} reasoning challenge.
\newblock \emph{arXiv preprint arXiv:1803.05457}, 2018.

\bibitem[Cobbe et~al.(2021)Cobbe, Kosaraju, Bavarian, Chen, Jun, Kaiser, Plappert, Tworek, Hilton, Nakano, Hesse, and Schulman]{cobbe2021gsm8k}
Karl Cobbe, Vineet Kosaraju, Mohammad Bavarian, Mark Chen, Heewoo Jun, Lukasz Kaiser, Matthias Plappert, Jerry Tworek, Jacob Hilton, Reiichiro Nakano, Christopher Hesse, and John Schulman.
\newblock Training verifiers to solve math word problems.
\newblock \emph{arXiv preprint arXiv:2110.14168}, 2021.

\bibitem[Dai et~al.(2024)Dai, Deng, Zhao, Xu, Gao, Chen, Li, Zeng, Yu, Wu, Xie, Li, Huang, Luo, Ruan, Sui, and Liang]{Dai2024DeepSeekMoETU}
Damai Dai, Chengqi Deng, Chenggang Zhao, R.~X. Xu, Huazuo Gao, Deli Chen, Jiashi Li, Wangding Zeng, Xingkai Yu, Y.~Wu, Zhenda Xie, Y.~K. Li, Panpan Huang, Fuli Luo, Chong Ruan, Zhifang Sui, and Wenfeng Liang.
\newblock Deepseekmoe: Towards ultimate expert specialization in mixture-of-experts language models.
\newblock \emph{ArXiv}, abs/2401.06066, 2024.
\newblock \url{https://api.semanticscholar.org/CorpusID:266933338}.

\bibitem[Douillard et~al.(2023)Douillard, Feng, Rusu, Chhaparia, Donchev, Kuncoro, Ranzato, Szlam, and Shen]{Douillard2023DiLoCoDL}
Arthur Douillard, Qixuang Feng, Andrei~A. Rusu, Rachita Chhaparia, Yani Donchev, Adhiguna Kuncoro, Marc'Aurelio Ranzato, Arthur Szlam, and Jiajun Shen.
\newblock Diloco: Distributed low-communication training of language models.
\newblock \emph{ArXiv}, abs/2311.08105, 2023.
\newblock \url{https://api.semanticscholar.org/CorpusID:265158012}.

\bibitem[Fedus et~al.(2022)Fedus, Zoph, and Shazeer]{fedus2022switch}
William Fedus, Barret Zoph, and Noam Shazeer.
\newblock Switch transformers: Scaling to trillion parameter models with simple and efficient sparsity.
\newblock \emph{The Journal of Machine Learning Research}, 23\penalty0 (1):\penalty0 5232--5270, 2022.

\bibitem[{Gemini Team}(2023)]{team2023gemini}
{Gemini Team}.
\newblock Gemini: a family of highly capable multimodal models.
\newblock \emph{arXiv preprint arXiv:2312.11805}, 2023.
\newblock Team, Gemini and Anil, Rohan and Borgeaud, Sebastian and Wu, Yonghui and Alayrac, Jean-Baptiste and Yu, Jiahui and Soricut, Radu and Schalkwyk, Johan and Dai, Andrew M and Hauth, Anja and others.

\bibitem[Gururangan et~al.(2021)Gururangan, Lewis, Holtzman, Smith, and Zettlemoyer]{Gururangan2021DEMixLD}
Suchin Gururangan, Michael Lewis, Ari Holtzman, Noah~A. Smith, and Luke Zettlemoyer.
\newblock Demix layers: Disentangling domains for modular language modeling.
\newblock In \emph{North American Chapter of the Association for Computational Linguistics}, 2021.
\newblock \url{https://api.semanticscholar.org/CorpusID:236976189}.

\bibitem[Gururangan et~al.(2023)Gururangan, Li, Lewis, Shi, Althoff, Smith, and Zettlemoyer]{gururangan2023scaling}
Suchin Gururangan, Margaret Li, Mike Lewis, Weijia Shi, Tim Althoff, Noah~A Smith, and Luke Zettlemoyer.
\newblock Scaling expert language models with unsupervised domain discovery.
\newblock \emph{arXiv preprint arXiv:2303.14177}, 2023.

\bibitem[Hendrycks et~al.(2021{\natexlab{a}})Hendrycks, Burns, Basart, Zou, Mazeika, Song, and Steinhardt]{DBLP:conf/iclr/HendrycksBBZMSS21}
Dan Hendrycks, Collin Burns, Steven Basart, Andy Zou, Mantas Mazeika, Dawn Song, and Jacob Steinhardt.
\newblock Measuring massive multitask language understanding.
\newblock In \emph{9th International Conference on Learning Representations, {ICLR} 2021, Virtual Event, Austria, May 3-7, 2021}. OpenReview.net, 2021{\natexlab{a}}.
\newblock \url{https://openreview.net/forum?id=d7KBjmI3GmQ}.

\bibitem[Hendrycks et~al.(2021{\natexlab{b}})Hendrycks, Burns, Kadavath, Arora, Basart, Tang, Song, and Steinhardt]{Hendrycks2021MeasuringMP}
Dan Hendrycks, Collin Burns, Saurav Kadavath, Akul Arora, Steven Basart, Eric Tang, Dawn~Xiaodong Song, and Jacob Steinhardt.
\newblock Measuring mathematical problem solving with the math dataset.
\newblock \emph{ArXiv}, abs/2103.03874, 2021{\natexlab{b}}.
\newblock \url{https://api.semanticscholar.org/CorpusID:232134851}.

\bibitem[Jacobs et~al.(1991)Jacobs, Jordan, Nowlan, and Hinton]{Jacobs1991AdaptiveMO}
Robert~A. Jacobs, Michael~I. Jordan, Steven~J. Nowlan, and Geoffrey~E. Hinton.
\newblock Adaptive mixtures of local experts.
\newblock \emph{Neural Computation}, 3:\penalty0 79--87, 1991.
\newblock \url{https://api.semanticscholar.org/CorpusID:572361}.

\bibitem[Jang et~al.(2016)Jang, Gu, and Poole]{jang2016categorical}
Eric Jang, Shixiang Gu, and Ben Poole.
\newblock Categorical reparameterization with gumbel-softmax.
\newblock \emph{arXiv preprint arXiv:1611.01144}, 2016.

\bibitem[Jiang et~al.(2024)Jiang, Sablayrolles, Roux, Mensch, Savary, Bamford, Chaplot, de~Las~Casas, Hanna, Bressand, Lengyel, Bour, Lample, Lavaud, Saulnier, Lachaux, Stock, Subramanian, Yang, Antoniak, Scao, Gervet, Lavril, Wang, Lacroix, and Sayed]{Jiang2024MixtralOE}
Albert~Q. Jiang, Alexandre Sablayrolles, Antoine Roux, Arthur Mensch, Blanche Savary, Chris Bamford, Devendra~Singh Chaplot, Diego de~Las~Casas, Emma~Bou Hanna, Florian Bressand, Gianna Lengyel, Guillaume Bour, Guillaume Lample, L'elio~Renard Lavaud, Lucile Saulnier, Marie-Anne Lachaux, Pierre Stock, Sandeep Subramanian, Sophia Yang, Szymon Antoniak, Teven~Le Scao, Th{\'e}ophile Gervet, Thibaut Lavril, Thomas Wang, Timoth{\'e}e Lacroix, and William~El Sayed.
\newblock Mixtral of experts.
\newblock \emph{ArXiv}, abs/2401.04088, 2024.
\newblock \url{https://api.semanticscholar.org/CorpusID:266844877}.

\bibitem[Joshi et~al.(2017)Joshi, Choi, Weld, and Zettlemoyer]{Joshi2017TriviaQAAL}
Mandar Joshi, Eunsol Choi, Daniel~S. Weld, and Luke Zettlemoyer.
\newblock Triviaqa: A large scale distantly supervised challenge dataset for reading comprehension.
\newblock \emph{ArXiv}, abs/1705.03551, 2017.
\newblock \url{https://api.semanticscholar.org/CorpusID:26501419}.

\bibitem[Komatsuzaki et~al.(2022)Komatsuzaki, Puigcerver, Lee-Thorp, Ruiz, Mustafa, Ainslie, Tay, Dehghani, and Houlsby]{Komatsuzaki2022SparseUT}
Aran Komatsuzaki, Joan Puigcerver, James Lee-Thorp, Carlos~Riquelme Ruiz, Basil Mustafa, Joshua Ainslie, Yi~Tay, Mostafa Dehghani, and Neil Houlsby.
\newblock Sparse upcycling: Training mixture-of-experts from dense checkpoints.
\newblock \emph{ArXiv}, abs/2212.05055, 2022.
\newblock \url{https://api.semanticscholar.org/CorpusID:254535822}.

\bibitem[Kwiatkowski et~al.(2019)Kwiatkowski, Palomaki, Redfield, Collins, Parikh, Alberti, Epstein, Polosukhin, Kelcey, Devlin, Lee, Toutanova, Jones, Chang, Dai, Uszkoreit, Le, and Petrov]{47761}
Tom Kwiatkowski, Jennimaria Palomaki, Olivia Redfield, Michael Collins, Ankur Parikh, Chris Alberti, Danielle Epstein, Illia Polosukhin, Matthew Kelcey, Jacob Devlin, Kenton Lee, Kristina~N. Toutanova, Llion Jones, Ming-Wei Chang, Andrew Dai, Jakob Uszkoreit, Quoc Le, and Slav Petrov.
\newblock Natural questions: a benchmark for question answering research.
\newblock \emph{Transactions of the Association of Computational Linguistics}, 2019.

\bibitem[Lange et~al.(2019)Lange, Aljundi, Masana, Parisot, Jia, Leonardis, Slabaugh, and Tuytelaars]{DeLange2019ACL}
Matthias~De Lange, Rahaf Aljundi, Marc Masana, Sarah Parisot, Xu~Jia, Ale{\v{s}} Leonardis, Gregory~G. Slabaugh, and Tinne Tuytelaars.
\newblock A continual learning survey: Defying forgetting in classification tasks.
\newblock \emph{IEEE Transactions on Pattern Analysis and Machine Intelligence}, 44:\penalty0 3366--3385, 2019.
\newblock \url{https://api.semanticscholar.org/CorpusID:218889912}.

\bibitem[Lewis et~al.(2021)Lewis, Bhosale, Dettmers, Goyal, and Zettlemoyer]{Lewis2021BASELS}
Mike Lewis, Shruti Bhosale, Tim Dettmers, Naman Goyal, and Luke Zettlemoyer.
\newblock Base layers: Simplifying training of large, sparse models.
\newblock In \emph{International Conference on Machine Learning}, 2021.
\newblock \url{https://api.semanticscholar.org/CorpusID:232428341}.

\bibitem[Li et~al.(2022{\natexlab{a}})Li, Gururangan, Dettmers, Lewis, Althoff, Smith, and Zettlemoyer]{Li2022BranchTrainMergeEP}
Margaret Li, Suchin Gururangan, Tim Dettmers, Mike Lewis, Tim Althoff, Noah~A. Smith, and Luke Zettlemoyer.
\newblock Branch-train-merge: Embarrassingly parallel training of expert language models.
\newblock \emph{ArXiv}, abs/2208.03306, 2022{\natexlab{a}}.
\newblock \url{https://api.semanticscholar.org/CorpusID:251371375}.

\bibitem[Li et~al.(2022{\natexlab{b}})Li, Choi, Chung, Kushman, Schrittwieser, Leblond, Tom, Eccles, Keeling, Gimeno, Lago, Hubert, Choy, de, d’Autume, Babuschkin, Chen, Huang, Welbl, Gowal, Alexey, Cherepanov, Molloy, Mankowitz, Robson, Kohli, de, Freitas, Kavukcuoglu, and Vinyals]{Li2022CompetitionlevelCG}
Yujia Li, David~H. Choi, Junyoung Chung, Nate Kushman, Julian Schrittwieser, R{\'e}mi Leblond, Tom, Eccles, James Keeling, Felix Gimeno, Agustin~Dal Lago, Thomas Hubert, Peter Choy, Cyprien de, Masson d’Autume, Igor Babuschkin, Xinyun Chen, Po-Sen Huang, Johannes Welbl, Sven Gowal, Alexey, Cherepanov, James Molloy, Daniel~Jaymin Mankowitz, Esme~Sutherland Robson, Pushmeet Kohli, Nando de, Freitas, Koray Kavukcuoglu, and Oriol Vinyals.
\newblock Competition-level code generation with alphacode.
\newblock \emph{Science}, 378:\penalty0 1092 -- 1097, 2022{\natexlab{b}}.
\newblock \url{https://api.semanticscholar.org/CorpusID:246527904}.

\bibitem[Ouyang et~al.(2022)Ouyang, Wu, Jiang, Almeida, Wainwright, Mishkin, Zhang, Agarwal, Slama, Ray, Schulman, Hilton, Kelton, Miller, Simens, Askell, Welinder, Christiano, Leike, and Lowe]{Ouyang2022TrainingLM}
Long Ouyang, Jeff Wu, Xu~Jiang, Diogo Almeida, Carroll~L. Wainwright, Pamela Mishkin, Chong Zhang, Sandhini Agarwal, Katarina Slama, Alex Ray, John Schulman, Jacob Hilton, Fraser Kelton, Luke~E. Miller, Maddie Simens, Amanda Askell, Peter Welinder, Paul~Francis Christiano, Jan Leike, and Ryan~J. Lowe.
\newblock Training language models to follow instructions with human feedback.
\newblock \emph{ArXiv}, abs/2203.02155, 2022.
\newblock \url{https://api.semanticscholar.org/CorpusID:246426909}.

\bibitem[Roller et~al.(2021)Roller, Sukhbaatar, Szlam, and Weston]{Roller2021HashLF}
Stephen Roller, Sainbayar Sukhbaatar, Arthur Szlam, and Jason Weston.
\newblock Hash layers for large sparse models.
\newblock In \emph{Neural Information Processing Systems}, 2021.
\newblock \url{https://api.semanticscholar.org/CorpusID:235367626}.

\bibitem[Rozi{\`e}re et~al.(2023)Rozi{\`e}re, Gehring, Gloeckle, Sootla, Gat, Tan, Adi, Liu, Remez, Rapin, Kozhevnikov, Evtimov, Bitton, Bhatt, Ferrer, Grattafiori, Xiong, D'efossez, Copet, Azhar, Touvron, Martin, Usunier, Scialom, and Synnaeve]{Rozire2023CodeLO}
Baptiste Rozi{\`e}re, Jonas Gehring, Fabian Gloeckle, Sten Sootla, Itai Gat, Xiaoqing Tan, Yossi Adi, Jingyu Liu, Tal Remez, J{\'e}r{\'e}my Rapin, Artyom Kozhevnikov, I.~Evtimov, Joanna Bitton, Manish~P Bhatt, Cristian~Cant{\'o}n Ferrer, Aaron Grattafiori, Wenhan Xiong, Alexandre D'efossez, Jade Copet, Faisal Azhar, Hugo Touvron, Louis Martin, Nicolas Usunier, Thomas Scialom, and Gabriel Synnaeve.
\newblock Code llama: Open foundation models for code.
\newblock \emph{ArXiv}, abs/2308.12950, 2023.
\newblock \url{https://api.semanticscholar.org/CorpusID:261100919}.

\bibitem[Rusu et~al.(2016)Rusu, Rabinowitz, Desjardins, Soyer, Kirkpatrick, Kavukcuoglu, Pascanu, and Hadsell]{Rusu2016ProgressiveNN}
Andrei~A. Rusu, Neil~C. Rabinowitz, Guillaume Desjardins, Hubert Soyer, James Kirkpatrick, Koray Kavukcuoglu, Razvan Pascanu, and Raia Hadsell.
\newblock Progressive neural networks.
\newblock \emph{ArXiv}, abs/1606.04671, 2016.
\newblock \url{https://api.semanticscholar.org/CorpusID:15350923}.

\bibitem[Sakaguchi et~al.(2021)Sakaguchi, Bras, Bhagavatula, and Choi]{sakaguchi2021winogrande}
Keisuke Sakaguchi, Ronan~Le Bras, Chandra Bhagavatula, and Yejin Choi.
\newblock Winogrande: An adversarial winograd schema challenge at scale.
\newblock \emph{Communications of the ACM}, 64\penalty0 (9):\penalty0 99--106, 2021.

\bibitem[Sap et~al.(2019)Sap, Rashkin, Chen, LeBras, and Choi]{sap2019socialiqa}
Maarten Sap, Hannah Rashkin, Derek Chen, Ronan LeBras, and Yejin Choi.
\newblock Socialiqa: Commonsense reasoning about social interactions.
\newblock \emph{arXiv preprint arXiv:1904.09728}, 2019.

\bibitem[Shao et~al.(2024)Shao, Wang, Zhu, Xu, Song, Zhang, Li, Wu, and Guo]{Shao2024DeepSeekMathPT}
Zhihong Shao, Peiyi Wang, Qihao Zhu, R.~X. Xu, Jun-Mei Song, Mingchuan Zhang, Y.~K. Li, Yu~Wu, and Daya Guo.
\newblock Deepseekmath: Pushing the limits of mathematical reasoning in open language models.
\newblock \emph{ArXiv}, abs/2402.03300, 2024.
\newblock \url{https://api.semanticscholar.org/CorpusID:267412607}.

\bibitem[Shazeer et~al.(2017)Shazeer, Mirhoseini, Maziarz, Davis, Le, Hinton, and Dean]{Shazeer2017OutrageouslyLN}
Noam~M. Shazeer, Azalia Mirhoseini, Krzysztof Maziarz, Andy Davis, Quoc~V. Le, Geoffrey~E. Hinton, and Jeff Dean.
\newblock Outrageously large neural networks: The sparsely-gated mixture-of-experts layer.
\newblock \emph{ArXiv}, abs/1701.06538, 2017.
\newblock \url{https://api.semanticscholar.org/CorpusID:12462234}.

\bibitem[Touvron et~al.(2023)Touvron, Martin, Stone, Albert, Almahairi, Babaei, Bashlykov, Batra, Bhargava, Bhosale, Bikel, Blecher, Ferrer, Chen, Cucurull, Esiobu, Fernandes, Fu, Fu, Fuller, Gao, Goswami, Goyal, Hartshorn, Hosseini, Hou, Inan, Kardas, Kerkez, Khabsa, Kloumann, Korenev, Koura, Lachaux, Lavril, Lee, Liskovich, Lu, Mao, Martinet, Mihaylov, Mishra, Molybog, Nie, Poulton, Reizenstein, Rungta, Saladi, Schelten, Silva, Smith, Subramanian, Tan, Tang, Taylor, Williams, Kuan, Xu, Yan, Zarov, Zhang, Fan, Kambadur, Narang, Rodriguez, Stojnic, Edunov, and Scialom]{touvron2023llama2}
Hugo Touvron, Louis Martin, Kevin Stone, Peter Albert, Amjad Almahairi, Yasmine Babaei, Nikolay Bashlykov, Soumya Batra, Prajjwal Bhargava, Shruti Bhosale, Dan Bikel, Lukas Blecher, Cristian~Canton Ferrer, Moya Chen, Guillem Cucurull, David Esiobu, Jude Fernandes, Jeremy Fu, Wenyin Fu, Brian Fuller, Cynthia Gao, Vedanuj Goswami, Naman Goyal, Anthony Hartshorn, Saghar Hosseini, Rui Hou, Hakan Inan, Marcin Kardas, Viktor Kerkez, Madian Khabsa, Isabel Kloumann, Artem Korenev, Punit~Singh Koura, Marie-Anne Lachaux, Thibaut Lavril, Jenya Lee, Diana Liskovich, Yinghai Lu, Yuning Mao, Xavier Martinet, Todor Mihaylov, Pushkar Mishra, Igor Molybog, Yixin Nie, Andrew Poulton, Jeremy Reizenstein, Rashi Rungta, Kalyan Saladi, Alan Schelten, Ruan Silva, Eric~Michael Smith, Ranjan Subramanian, Xiaoqing~Ellen Tan, Binh Tang, Ross Taylor, Adina Williams, Jian~Xiang Kuan, Puxin Xu, Zheng Yan, Iliyan Zarov, Yuchen Zhang, Angela Fan, Melanie Kambadur, Sharan Narang, Aurelien Rodriguez, Robert Stojnic, Sergey Edunov, and Thomas
  Scialom.
\newblock Llama 2: Open foundation and fine-tuned chat models, 2023.

\bibitem[Wortsman et~al.(2022)Wortsman, Ilharco, Gadre, Roelofs, Gontijo-Lopes, Morcos, Namkoong, Farhadi, Carmon, Kornblith, and Schmidt]{Wortsman2022ModelSA}
Mitchell Wortsman, Gabriel Ilharco, Samir~Yitzhak Gadre, Rebecca Roelofs, Raphael Gontijo-Lopes, Ari~S. Morcos, Hongseok Namkoong, Ali Farhadi, Yair Carmon, Simon Kornblith, and Ludwig Schmidt.
\newblock Model soups: averaging weights of multiple fine-tuned models improves accuracy without increasing inference time.
\newblock \emph{ArXiv}, abs/2203.05482, 2022.
\newblock \url{https://api.semanticscholar.org/CorpusID:247362886}.

\bibitem[Xue et~al.(2024)Xue, Zheng, Fu, Ni, Zheng, Zhou, and You]{xue2024openmoe}
Fuzhao Xue, Zian Zheng, Yao Fu, Jinjie Ni, Zangwei Zheng, Wangchunshu Zhou, and Yang You.
\newblock Openmoe: An early effort on open mixture-of-experts language models.
\newblock \emph{arXiv preprint arXiv:2402.01739}, 2024.

\bibitem[Zhang et~al.(2015)Zhang, Choromanska, and LeCun]{NIPS2015_d18f655c}
Sixin Zhang, Anna~E Choromanska, and Yann LeCun.
\newblock Deep learning with elastic averaging sgd.
\newblock In C.~Cortes, N.~Lawrence, D.~Lee, M.~Sugiyama, and R.~Garnett, editors, \emph{Advances in Neural Information Processing Systems}, volume~28. Curran Associates, Inc., 2015.
\newblock \url{https://proceedings.neurips.cc/paper_files/paper/2015/file/d18f655c3fce66ca401d5f38b48c89af-Paper.pdf}.

\bibitem[Zhang et~al.(2022)Zhang, Roller, Goyal, Artetxe, Chen, Chen, Dewan, Diab, Li, Lin, Mihaylov, Ott, Shleifer, Shuster, Simig, Koura, Sridhar, Wang, and Zettlemoyer]{Zhang2022OPTOP}
Susan Zhang, Stephen Roller, Naman Goyal, Mikel Artetxe, Moya Chen, Shuohui Chen, Christopher Dewan, Mona~T. Diab, Xian Li, Xi~Victoria Lin, Todor Mihaylov, Myle Ott, Sam Shleifer, Kurt Shuster, Daniel Simig, Punit~Singh Koura, Anjali Sridhar, Tianlu Wang, and Luke Zettlemoyer.
\newblock Opt: Open pre-trained transformer language models.
\newblock \emph{ArXiv}, abs/2205.01068, 2022.
\newblock \url{https://api.semanticscholar.org/CorpusID:248496292}.

\bibitem[Zhao et~al.(2024)Zhao, Zhang, Zhang, Gui, and Huang]{zhao2024llama}
Jun Zhao, Zhihao Zhang, Qi~Zhang, Tao Gui, and Xuanjing Huang.
\newblock Llama beyond english: An empirical study on language capability transfer.
\newblock \emph{arXiv preprint arXiv:2401.01055}, 2024.

\end{thebibliography}
